\documentclass[runningheads]{llncs}

\usepackage[mobile]{eccv}

\usepackage{eccvabbrv}

\usepackage{graphicx}
\usepackage{booktabs}

\usepackage[accsupp]{axessibility}  

\usepackage{hyperref}
\usepackage{orcidlink}
\usepackage{booktabs}
\usepackage{multirow}
\usepackage[normalem]{ulem}
\useunder{\uline}{\ul}{}

\begin{document}

\title{Identifying and Resolving Pitfalls of Knowledge-Based VQA Benchmarks: Auditing, Repairing, and Augmenting}

\titlerunning{KB-VQA: Auditing, Repairing, and Augmenting}

\author{Qian Ma\inst{1}\orcidlink{0009-0006-5194-7642} \and
S M Rayeed\inst{1}\orcidlink{0000-0001-7173-3429} \and
Charles V. Stewart\inst{1}\orcidlink{0000-0001-6532-6675} \and 
Qiong Wu\inst{2}\orcidlink{0000-0001-7724-8221} \and
Yao Ma\inst{1}\orcidlink{0000-0002-4985-8724}
}

\authorrunning{Q. Ma et al.}

\institute{Rensselaer Polytechnic Institute, Troy NY 12047, USA \and
AT\&T Chief Data Office, Bedminster NJ 07921, USA \\
\email{\{maq5,rayees,stewart,may13\}@rpi.edu}, qw6547@att.com
} 

\maketitle

\begin{abstract}

Knowledge-Based Visual Question Answering (KB-VQA) aims to evaluate whether Visual Language Models (VLMs) can retrieve, ground, and reason over external structured knowledge beyond visual evidence. In practice, answer accuracy is widely adopted as the primary evaluation metric, implicitly treating correctness as a proxy for knowledge-grounded reasoning. However, for existing KB-VQA benchmarks, this proxy relies on critical assumptions that are often overlooked and rendered unreliable by benchmark issues: annotated answer must be derivable from the associated knowledge base, question must be well-posed with sufficient constraints, and visual setting must meaningfully require grounded disambiguation.
In this work, we show that these assumptions are systematically violated in existing KB-VQA benchmarks. Our audit reveals substantial instances with missing or contradicted answers and underspecified questions that render accuracy a misleading metric. Furthermore, we find that existing datasets rely on visually trivial, single-entity scenes that bypass the need for sophisticated visual-to-knowledge mapping. We demonstrate that even with controlled architectures, these flaws lead to distorted model rankings and overestimations of reasoning capabilities.
To address this, we introduce (1) a principled audit-and-repair protocol that restores answer derivability and question clarity, and (2) a controlled multi-entity augmentation protocol that introduces visual ambiguity to challenge initial retrieval and grounded reasoning. Re-evaluation under corrected and augmented settings yields markedly different performance trends. Our findings call for rethinking evaluation protocols and designing more interaction-aware KB-VQA benchmarks that prioritize verifiable reasoning over simple matching.
\setcounter{footnote}{0}
\footnote{The datasets and code are available in \url{https://github.com/VAN-QIAN/ECCV26-ARA}.\\
Work was initiated when Qian was an intern at AT\&T CDO.
Qiong Wu and Yao Ma are co-corresponding authors.}
\end{abstract}
\section{Introduction}\label{sec:intro}
Recent Vision-Language Models (VLMs)~\cite{achiam2023gpt,bai2025qwen2,wang2025internvl3,zhang2024vision} have demonstrated strong performance on a variety of visual reasoning tasks, especially Visual Question Answering (VQA) settings that require aligning image content with language understanding~\cite{antol2015vqa,goyal2017making,kim2025visual,kuang2025natural}.
Despite such effectiveness, they are still struggling to address tasks where answer is beyond the visual part like Knowledge-Based Visual Question Answering (KB-VQA).
KB-VQA~\cite{InfoSeek,EVQA} is designed to evaluate whether a model can answer image-grounded questions correctly by retrieving and reasoning over an external, controlled knowledge base, rather than relying only on parametric memory~\cite{InfoSeek,deng2025comprehensive}, where accuracy is used to assess such knowledge-grounded capability.

{Emerging efforts are focusing on achieving better answer accuracy by developing re-ranker modules to select the most relevant and informative evidence segment for the answer generation~\cite{IBA,Echosight}, improving the model's inherent capability to use all retrieved knowledge~\cite{CoMeM,ReflectiVA,CCVQA,reag} or empowering VLMs to invoke external tools to retrieve more relevant evidence~\cite{Wiki-PRF}. Significantly increasing efforts and resources are being invested to achieve better accuracy score.}

However, in representative KB-VQA benchmarks such as InfoSeek and E-VQA~\cite{EVQA,InfoSeek}, we observe recurring dataset issues (Section~\ref{sec:issues}) that can make answer accuracy an unreliable indicator of the intended knowledge-grounded reasoning capability.
\textbf{1. Answer-evidence misalignment.}
A non-trivial portion of questions have annotated answers that are not supported by the provided knowledge resource (\eg missing or contradicted).
{As shown in Figure~\ref{fig:annotation_issues}, when answering the sea level of the memorial park, the desired answer is `10 foot' while it's missing in the benchmark-provided knowledge base.} 
\textbf{2. Underspecified questions and answer scope.} 
Many questions do not specify the intended answer granularity, so multiple answers can be reasonable.
For example, for question ``When is the mating period of this bird?'', the evidence may mention a season (Spring), specific months (March-April), or a life-stage (around 1-year old).
When several evidence-supported answers are plausible, accuracy becomes sensitive to the single particular annotated answer, and can therefore understate a model's knowledge-grounded capability.
\textbf{3. Visually simplified scenes vs.\ real-world multi-modal queries.}
Existing KB-VQA benchmarks~\cite{InfoSeek,EVQA} present visually simplified scenes dominated by a single salient entity, so questions can often be answered using coarse global cues and retrieval may succeed without explicit grounding to the intended target~\cite{sun2024eva,radford2021learning}.
In realistic scenarios, however, users often issue more complex multi-modal queries that require text-image interaction \ie the query text specifies which entity is relevant (\eg, via spatial relations, relational descriptions, \etc), and the system should localize candidate entities and resolve the intended entity before KB retrieval and reasoning.
For example, a query may refer to ``the fish on the right'' where multiple plausible entities coexist and global image similarity alone is insufficient.

These issues matter because they decouple benchmark scores from the capability KB-VQA aims to measure. Under such conditions, high answer accuracy can be driven by annotation artifacts, dataset biases, or shortcut strategies, rather than faithful knowledge-grounded reasoning.

To address these limitations, we propose a unified fixing protocol and augmentation framework. The fixing protocol enforces answer derivability and question clarity via evidence verification, answer-source consistency checks, and targeted question repair (Section~\ref{sec:fixing-pipeline}). Building on the repaired benchmark, we further introduce a controlled augmentation pipeline that injects visual distractors while preserving core QA semantics, creating more challenging multi-entity settings for grounded retrieval and reasoning (Section~\ref{sec:aug-pipeline}).

Re-evaluation with representative KB-VQA baselines~\cite{IBA,Echosight,Wiki-PRF,ReflectiVA,CoMeM} reveals that correcting dataset validity alone can materially change measured performance and method ranking, while augmentation causes substantial drops in retrieval recall and end-to-end QA accuracy (Section~\ref{sec:Experiments}). These findings suggest that current KB-VQA evaluation can be overly optimistic, and that robust progress requires both annotation-level validity and explicit grounding difficulty.
\section{Related Works}\label{sec:related-works}
\subsection{Existing KB-VQA Datasets}\label{subsec:existing-kbvqa}
The transition from general VQA~\cite{antol2015vqa,goyal2017making} to KB-VQA~\cite{EVQA,InfoSeek} marks a shift from internal visual perception to the integration of open-world knowledge. While early benchmarks like OK-VQA~\cite{marino2019ok} and FVQA~\cite{FVQA} focused on general common-sense reasoning, modern datasets have scaled toward massive, external knowledge sources to simulate more complex grounding. Concurrent works such as \textbf{InfoSeek}~\cite{InfoSeek} and \textbf{E-VQA}~\cite{EVQA} are two primary examples of this large-scale shift. InfoSeek introduces over 1.3 million samples to simulate real-world intent; however, it suffers from a paradigm flaw where the knowledge base often lacks the specific facts used during human annotation, leading to misalignment and evaluation distortion. E-VQA attempts to address this by providing a controlled corpus of 2 million Wikipedia pages with section-level evidence. Despite these improvements, both datasets rely heavily on template-based question generation, which often lacks the linguistic variety found in natural queries. 
Beyond these two datasets, there is a following work SK-VQA~\cite{su2025skvqa} that claims context comprehension capability is important to achieve KB-VQA task.
Therefore, they constructed a synthetic dataset to generate `pseudo' wiki articles to explicitly train the model to seek information among it.
Beyond task formulation, prior KB-VQA benchmarks~\cite{InfoSeek,EVQA} also differ in how their knowledge sources, annotations, and question templates are constructed.
These design choices can introduce recurring issues such as mismatches between annotated answers and retrievable evidence, underspecified templates that admit multiple plausible answers, and visually simplified scenes that weaken the need for grounded disambiguation.
Our work builds on these benchmarks and provides a structured characterization of such issues and their implications for evaluation.

\subsection{MM-RAG Solutions for KB-VQA}\label{subsec:mm-rag-solutions}

Since the answers of KB-VQA questions are beyond the visual part and external knowledge is required, most emerging solutions for KB-VQA tasks can be categorized into Multi-modal Retrieval-Augmented Generation (MM-RAG) frameworks to effectively acquire the external knowledge base according to the multi-modal queries. 
These solutions generally focus on how to selectively utilize retrieved content to respond to queries and can be categorized into two streams.

\noindent \textbf{1. Re-ranking based methods} mainly focus on picking the most relevant and informative part of all initial-retrieved information \eg the section corresponding to target question of all retrieved Wikipedia articles. 
\textbf{IBA}~\cite{IBA} addresses this by explicitly decoupling the workflow to ground the target entity by doing identification before reasoning to answer. It leverages the zero-shot recognition capabilities of MLLMs to identify candidate entities before employing a lightweight text-re-ranker for precise evidence selection. 
This modular workflow re-design approach contrasts with framework \textbf{EchoSight}~\cite{Echosight}, which trains a customized multi-modal re-ranker based on Q-Former\cite{li2023blip}.

\noindent \textbf{2. Aggregation/Filtering-based} methods focusing on digesting all retrieved information to implicitly provide the evidence for downstream answer generation  instead of explicitly selecting the evidence.
\textbf{ReflectiVA}\cite{ReflectiVA} introduces specialized self-reflective tokens that allow the re-trained MLLM to autonomously determine the necessity of each retrieved section. 
All sections considered as relevant will be aggregated to support the answer generation.
\textbf{CoMeM}~\cite{CoMeM} proposed to condense all retrieved articles into the embeddings and concatenate with query embeddings to support the answer generation as model memory. 
In \textbf{Wiki-PRF}~\cite{Wiki-PRF}, reinforcement learning is introduced to empower MLLMs to conduct multiple retrievals with different tools and filtering all retrieved results to support the final answer generation.

\section{Issues in Existing KB-VQA Benchmarks}\label{sec:issues}

Answer accuracy is widely reported on KB-VQA benchmarks, but it can be unreliable when benchmark instances violate basic validity requirements.
Across E-VQA~\cite{EVQA} and InfoSeek~\cite{InfoSeek}, we observe three recurring dataset issues that may directly distort evaluation: (i) \textbf{answer-evidence misalignment} that creates unavoidable false negatives, (ii) \textbf{underspecified questions} where multiple answers are defensible and scores depend on the single desired annotated answer, and (iii) \textbf{visually simplified scenes} where shortcut retrieval can succeed without grounded disambiguation.
Together, these issues can decouple benchmark scores from the knowledge-grounded retrieval and reasoning capability KB-VQA is intended to measure.
We use three assumptions, \emph{(A) Answer Derivability,} \emph{(B) Question Well-Posedness,} and \emph{(C) Grounded Disambiguation Requirement} as an organizing lens, and document representative violations under this framework.

\subsection{Answer-Evidence Misalignment}
Assumption (A) requires that each annotated answer be derivable from the associated KB evidence. 
We observe two recurring violations: 
\emph{(a) Unsupported answers}, where the answer is absent from the KB (\eg InfoSeek QID 23994 annotates ``10 foot'' for sea level of Wright Brothers National Memorial while corresponding evidence article doesn't contain any relevant information), and 
\emph{(b) Contradictory answers}, where annotation conflicts with evidence (\eg InfoSeek QID 5411 annotates ``1,302'' kilogram as Mass of McLaren car, but evidence states ``It weighs 1,301 kg'' such a single value won't match the desired range).

When derivability fails, evaluation produces unavoidable false negatives under the benchmark-provided KB that even perfect retrieval and correct reasoning can be scored as wrong because the target annotation is not grounded in the retrievable evidence (Figure~\ref{fig:annotation_issues}).
Using an initial verification pass with Qwen3-30B-A3B~\cite{yang2025qwen3} to scan each target entity page and judge support and derivability, 
we find unsupported annotations in about 1\% of E-VQA and 22\% of InfoSeek.
This implies that a non-trivial fraction of InfoSeek is unanswerable under its own provided KB, capping accuracy by construction.
This failure is largely structural in InfoSeek, where QA pairs come from WikiData~\cite{vrandevcic2014wikidata} knowledge graph triplets but evaluation uses Wikipedia text, creating a systematic cross-source mismatch.

\subsection{Underspecified Questions}
Assumption (B) requires each question to be sufficiently constrained so that the annotated answer is unique under the given KB.
We identify three frequent ambiguity patterns:

\noindent \emph{(1) Missing attribute constraint} \eg For the question ``How big can this plant become?'' without specifying height or diameter.

\noindent \emph{(2) Missing temporal scope} \eg ``When is the mating period?'' without phase or time granularity, then the answer could be a season(spring) or month(September) during the year or during the species' life-cycle(such as `1-year old').

\noindent \emph{(3) Missing spatial reference or granularity} (\eg For the same question ``In which country or region does this animal live?'', the desired habitat answers could be region-level(North America) or country-level(United States) ).

Using these tags, our audit shows that 59\% of E-VQA and 47\% of InfoSeek questions are ambiguous: 
E-VQA has 21.5\% attribute, 27.5\% spatial, and 10\% temporal omissions; 
InfoSeek shows a similar pattern (17.3\%, 30.3\%, 2\%).
Under such under-specification, models may produce semantically valid alternatives but still be penalized as wrong. 
Therefore, accuracy becomes sensitive to annotation preference rather than reasoning correctness. 
These ambiguities mainly stem from template-based generation~\cite{EVQA} and automatic KG-to-question mapping~\cite{InfoSeek}, where key qualifiers are dropped.

\subsection{Single-Entity Shortcut and Missing Grounded Disambiguation}
Assumption (C) requires that success depends on grounded disambiguation to respond to complex multi-modal user queries under realistic scenarios. 
Ideally, KB-VQA involves four stages: entity localization, text-image grounding, entity disambiguation, and KB retrieval/reasoning.
In realistic queries (\eg ``the fish on the left''), the model should first identify which entity is being referenced. 
However, existing KB-VQA benchmarks usually contain one dominant entity per image and evaluate against a known target identifier.
Because the target is effectively fixed per instance, global retrieval can succeed without verifying which entity the question refers to.
With this design, global image embeddings can often retrieve the target directly, collapsing localization, grounding and disambiguation into a shortcut.
As a result, high answer accuracy does not necessarily imply robust grounding ability. 
The metric can overestimate capability because shortcut retrieval is sufficient in many samples. 
This issue is driven by entity-centric curation, target-anchored evaluation protocols, and reliance on global embedding retrieval~\cite{EVQA,InfoSeek}.
To restore this missing requirement, Section~\ref{sec:aug-pipeline} introduces controlled multi-entity augmentation so that localization, grounding, and disambiguation become necessary for success.
\section{Proposed Fixing Protocol}\label{sec:fixing-pipeline}

To address recurring benchmark issues such as \textbf{answer-evidence mismatch} and \textbf{underspecified} questions (Section~\ref{sec:issues}), we propose a dataset-level audit-and-repair protocol applicable to KB-VQA benchmarks.
It explicitly enforces the two assumptions identified in Section~\ref{sec:issues}:
\emph{Assumption A (Answer Derivability):} the annotated answer must be supported by and derivable from the external knowledge base;
\emph{Assumption B (Question Well-Posedness):} the question must provide sufficient constraints to uniquely determine the annotated answer.
Given a KB-VQA instance (image, QA pair, target entity, and KB evidence snapshot),
the pipeline outputs either a repaired instance or a filtered-out instance.
It consists of four cascaded stages: evidence verification, answer-derivability auditing,
question-constraint repair, and final consistency validation.

\subsection{General Four-Stage Protocol}

\begin{figure*}[!htbp]
    \centering
    \includegraphics[width=0.85\linewidth]{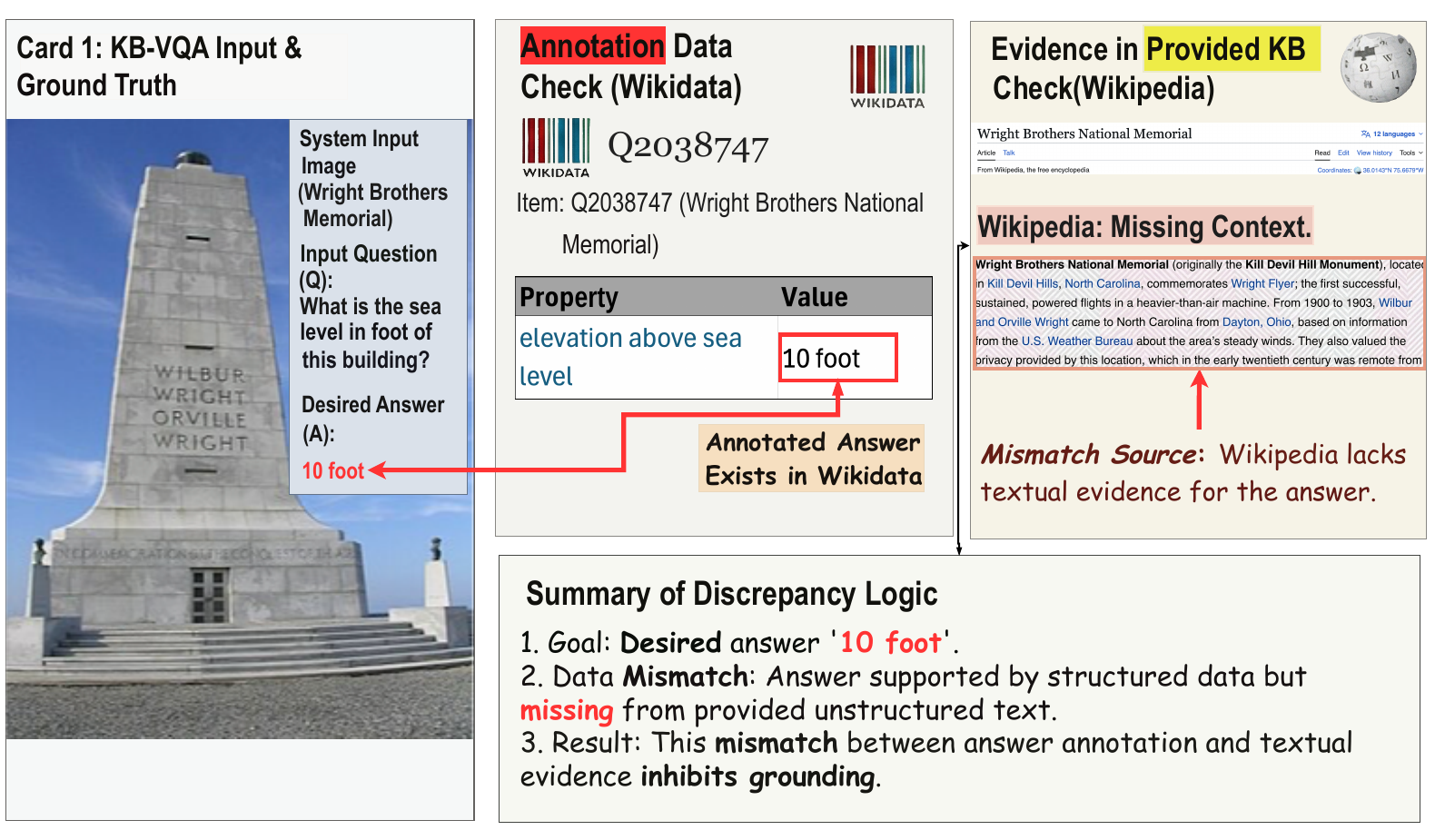}
    \caption{Qualitative example from InfoSeek (QID:149). InfoSeek~\cite{InfoSeek} selects answers from Wikidata~\cite{vrandevcic2014wikidata} triples and converts them into QA pairs, while the evaluation KB consists of Wikipedia articles. This cross-source construction can yield cases where the provided KB does not support the annotated answer, confounding score interpretation even under perfect retrieval.}
    \label{fig:annotation_issues}
\end{figure*}

\noindent \textbf{Step 1: Evidence Verification.}
We first verify whether the evidence referenced by each QA pair exists in the KB snapshot and
whether it contains support for the annotated answer under the target entity context.
Instances are categorized as \textbf{Supported \& Matched}, \textbf{Supported but Mismatched}, and \textbf{Unsupported}.
Since InfoSeek constructs QA pairs from Wikidata while using Wikipedia articles as the evaluation KB,
we scan the target-entity page section by section and apply two independent verifiers (Qwen3-30B-A3B~\cite{yang2025qwen3} and DeepSeek-v3.2~\cite{liu2025deepseek})
to mitigate erroneous filtering due to cross-source mismatch. Only instances that are consistently classified as \textbf{Unsupported} by both verifiers are removed (Figure~\ref{fig:annotation_issues}).
The verification is constrained to checking whether the target-entity evidence supports, contradicts, or lacks the annotated answer, rather than making an open-ended judgment.
The evidence context is localized to sections, with mean/95th-percentile lengths of 868/2360 tokens for E-VQA and 877/2545 tokens for InfoSeek.
For E-VQA, where evidence segments are provided during dataset construction, we directly examine the referenced segment and revisit cases with answer--evidence mismatch.

\noindent \textbf{Step 2: Answer-Derivability Auditing and Calibration.}
Given verified evidence and the original question, we test whether the annotated answer is logically derivable.
For any instances tagged as \textbf{Supported but Mismatched}, if the answer contradicts evidence, we apply evidence-grounded answer correction.
DeepSeek-v3.2 is not used as an answer source. 
It only rewrites a mismatched annotation to the value explicitly supported by verified evidence.
If no evidence-supported value can be derived, the instance is removed.
For InfoSeek, this step frequently corrects answer-side mismatch introduced by cross-source construction.
For E-VQA~\cite{EVQA}, answer-evidence contradictions are much less frequent, consistent with its construction~\cite{changpinyo2022all} where the answer is selected from the provided evidence article.

\noindent \textbf{Step 3: Question-Constraint Repair.}
We identify whether the question is sufficiently constrained for unique answering.
We repair ambiguous questions with minimal edits using three tags:
missing attribute constraints, missing temporal scope, and missing spatial reference.
Edits preserve the original intent and evidence dependency.
For InfoSeek, the dropped qualifiers from KG-to-question mapping are restored by adding missing facets.
For E-VQA, the original template-generated questions to fit super-category are revised mainly when templates omit disambiguating qualifiers to the fine-grained entity.

\noindent \textbf{Step 4: Leakage and Global Consistency Validation.}
After repair, we run a final pass to ensure that: (i) the question does not leak the answer,
(ii) the answer remains evidence-supported, and (iii) the revised question yields a unique answer.
Any leakage-inducing edits are rolled back and rewritten conservatively.

\subsection{Fixing Outcomes and Human Evaluation}\label{sec:fixed-outcome}

Applying the same four-stage protocol to both datasets yields different failure profiles but a unified corrected setting.
For InfoSeek, we retain 58,285 out of 71,335 instances (81.7\%) after filtering and repair.
We construct an entity-unique subset of 1,924 questions(consisting 1604 String, 223 Numerical and 97 Time questions) for controlled evaluation \ie for instances targeting on the same target entity with same query text but different query images, we only keep one.
For E-VQA, answer-side conflicts are rarer and most edits happen in question constraints. Hence the fixed evaluation split remains 4,750 questions.
For human evaluation, we expand the review to 10\% of the fixed evaluation splits, covering 475 E-VQA and 200 InfoSeek instances.
Following existing work~\cite{su2025skvqa}, two PhD-student-level annotators answer each fixed query using only the oracle evidence of the target entity.
Human accuracy remains high: $92.9 \pm 1.1$ for fixed E-VQA and $91.5 \pm 2.1$ for fixed InfoSeek.
More qualitative examples of fixing outcomes and human-evaluation cases are provided in Appendix~\ref{app:fixing-examples} and Appendix~\ref{app:fixing-human-eval}.
These repaired datasets are used as the \emph{fixed} versions and original datasets with the same index/identifier are used as the \emph{unfixed} in Section~\ref{sec:Experiments}.
\section{Proposed Augmentation Protocol}\label{sec:aug-pipeline}

As discussed in Section~\ref{sec:issues}, a gap remains between current KB-VQA benchmark settings and real-world multi-modal queries, where users often must specify the target entity through text alone under multi-entity ambiguity (\eg, ``the fish on the right'').
In existing benchmarks, however, images are frequently dominated by a single salient entity, so retrieval can succeed via coarse global image similarity overlooking \emph{(C) Grounded Disambiguation Requirement}.

\subsection{General augmentation protocols}
To close the evaluation gap in Section~\ref{sec:issues}, we augment each instance by adding a single distractor entity alongside the original target (anchor).
We preserve the annotated answer by construction and keep the external knowledge source unchanged, so performance changes reflect increased visual ambiguity rather than knowledge or annotation shifts.
Each augmented image contains exactly one anchor and one distractor, where the distractor is drawn from either the same semantic category (intra-category) or a different one (inter-category), forcing retrieval to rely on grounded disambiguation rather than coarse global similarity.

\noindent \textbf{Intra-Category Distraction}
We spatially concatenate the anchor image with a single distractor from the same semantic category, placed in a different region (\eg, left vs.\ right).
We minimally revise the question to reference the anchor region or a distinguishing attribute (\eg, ``the fish on the left''), while keeping the annotated answer unchanged.
This setting isolates fine-grained intra-category ambiguity and tests whether models ground textual constraints in localized visual evidence rather than relying on coarse global embeddings.

\noindent \textbf{Inter-Category Distraction}
We pair the anchor entity with a single distractor from a different semantic category (\eg, a landmark with an animal, or a plant with a man-made object).
Although visually distinct, the distractor can interrupting global image representations, requiring models to integrate the question with localized evidence to infer the intended referent.
The question and answer remain unchanged, allowing us to test whether retrieval is overly sensitive to irrelevant visual content instead of grounding target entity under textual constraints.

\subsection{Augmentation Outcomes and Human Evaluation}
\begin{figure}
    \centering
    \includegraphics[width=0.95\linewidth]{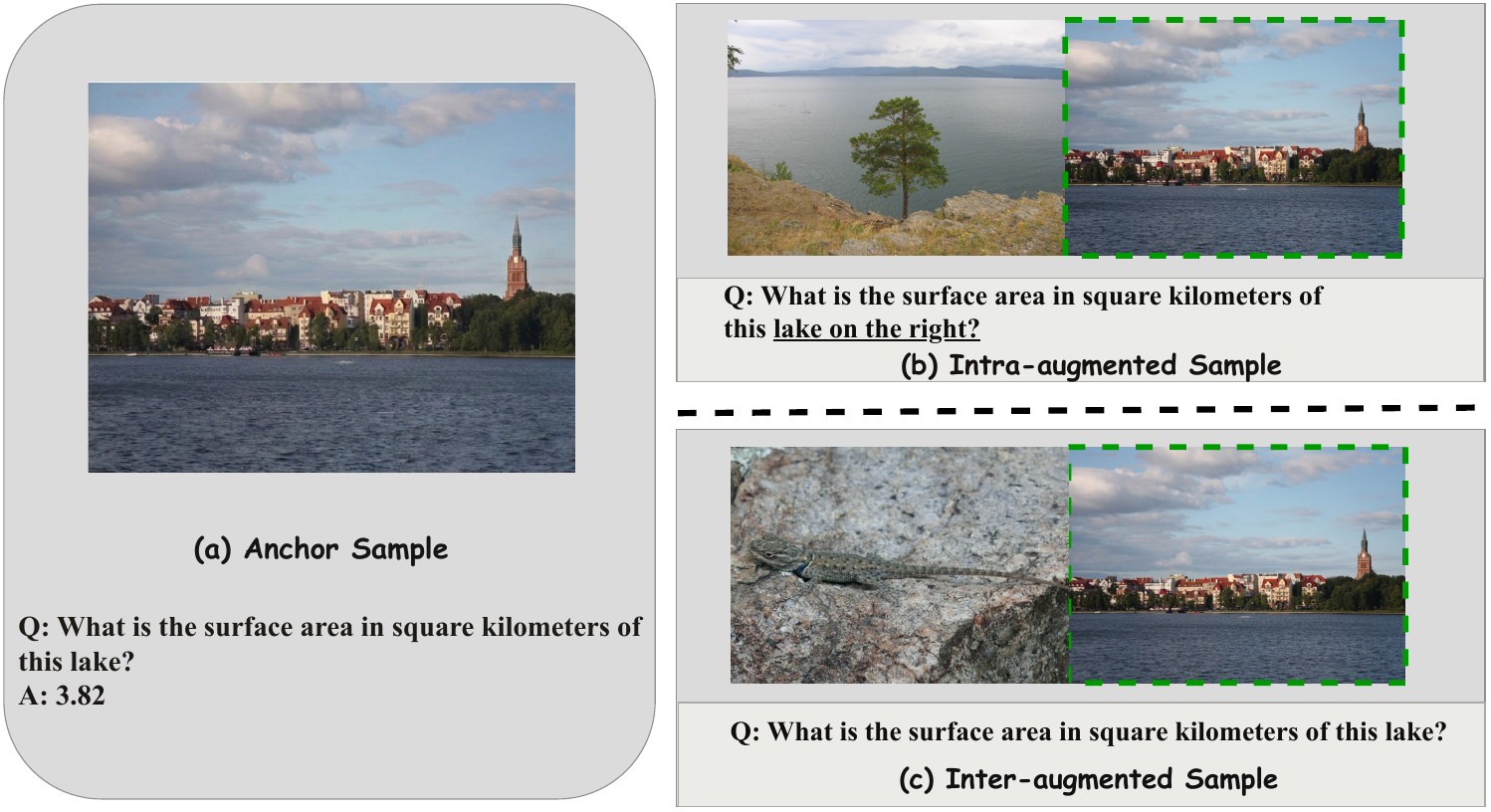}
    \caption{Qualitative example of our augmentation protocols (Section~\ref{sec:aug-pipeline}) on \emph{E\l{}k Lake}. All augmented variants keep the anchor answer in (a). (b) Intra-augmentation adds a minimal spatial cue (``on the right'') to preserve the original intent while inserting an additional lake image (\emph{Lake Turgoyak}) on the left. (c) Inter-augmentation inserts a visually distinct distractor (\emph{Sceloporus jarrovii}) while keeping the question unchanged.}
    \label{fig:augmentation-example1}
\end{figure}

To enable direct comparisons, we exclude questions that would remain ambiguous after intra-category augmentation (\eg, those referring to ``the object'') from \emph{both} splits.
As shown in Figure~\ref{fig:augmentation-example1}, intra-category augmentation adds a minimal cue (typically spatial) to specify the anchor entity, whereas inter-category augmentation keeps the question unchanged.
With answers preserved by construction, performance changes reflect increased grounded disambiguation demands rather than knowledge or annotation shifts.
We generate 1,604 (InfoSeek) and 3,871 (E-VQA) augmented variants, with additional examples in Appendix~\ref{app:augmentation-examples}.
For quality assurance, we sample 100 intra- and 100 inter-category instances per dataset.
Two annotators answer each augmented query given the anchor evidence and optionally report evident flaws~\cite{su2025skvqa}.
Human accuracy is $87.5\pm2.1\%$/ $96.0\pm1.4\%$ (E-VQA intra/inter) and $86.0\pm1.4\%$/ $89.5\pm3.5\%$ (InfoSeek intra/inter). More details are in Appendix~\ref{app:Augmentation-human-eval}.

\section{Experiments}\label{sec:Experiments}

\subsection{Experimental Setup}

\noindent \textbf{Datasets.}
We evaluate KB-VQA baselines on the following aligned dataset variants: 
(1) the original \emph{unfixed} benchmark, 
(2) a \emph{fixed} version that enforces Assumption~A/B via the pipeline in Section~\ref{sec:fixing-pipeline},
and (3) the~\emph{Intra} and~\emph{Inter} augmented versions that enforces the grounded-disambiguation requirement via the pipeline in Section~\ref{sec:aug-pipeline}.
Overall, these controlled comparisons tie each analysis directly to the assumptions in Section~\ref{sec:issues} and validate whether fixing and augmentation restore the intended evaluation conditions.

\noindent \textbf{External Knowledge Base.}
E-VQA provides a controlled knowledge base of 2 million Wikipedia articles with associated images.
Following prior work~\cite{Echosight}, we focus on the single-hop setting.
For InfoSeek, following existing baselines~\cite{Wiki-PRF,CoRe-MMRAG,ReflectiVA}, we adopt the 100K-article knowledge base released by Yan and Xie~\cite{Echosight} and follow the same initial retrieval settings using EVA-CLIP-8B~\cite{sun2024eva}.

\noindent \textbf{Baselines.}
We benchmark five representative KB-VQA methods, grouped by how retrieved evidence is utilized.
\textit{Re-ranking-based methods} (IBA~\cite{IBA}, EchoSight~\cite{Echosight}) explicitly re-rank retrieved evidence and generate the final answer conditioned on the top-ranked section, using either a tailored workflow paradigm or a trained multimodal re-ranker.
\textit{Aggregation/Filtering-based methods} (ReflectiVA~\cite{ReflectiVA}, CoMeM\cite{CoMeM}, Wiki-PRF~\cite{Wiki-PRF}) instead aggregate evidence across multiple sections or retrieval rounds and filter irrelevant content before answer generation.
ReflectiVA assigns section-level special relevant tokens and aggregates the selected evidence.
CoMeM encodes retrieved articles into memory representations that are consumed by an MLLM.
Wiki-PRF performs iterative retrieval with auxiliary tools (\eg grounding and captioning) and filters retrieved content across rounds to serve as evidence for answer generation.

\noindent \textbf{Evaluation Metrics.}
We report both end-to-end QA accuracy and retrieval-related metrics.
For answer accuracy, we evaluate open-ended outputs using BEM~\cite{zhang2019bertscore} on E-VQA.
For InfoSeek, following prior practice~\cite{Echosight,ReflectiVA,Wiki-PRF}, we use VQA accuracy~\cite{goyal2017making,marino2019ok,Wiki-PRF}.
For retrieval, we report Recall@K, indicating whether the correct target-entity article appears among the top-$K$ retrieved results.
For methods that include an explicit re-ranking stage~\cite{Echosight} or multi-round retrieval~\cite{Wiki-PRF}, we additionally report \emph{post-retrieval} performance, measuring whether the final evidence used for answer generation (\eg the top-1 re-ranked section or the aggregated evidence pool) contains the correct target entity.

\noindent \textbf{Implementation Details.}
For all baselines~\cite{Echosight,ReflectiVA,Wiki-PRF,CoMeM}, we use officially released checkpoints and prompts with unified KB and retrieval settings.
For IBA~\cite{IBA} component selection, we use Qwen2.5-VL-7B-Instruct~\cite{bai2025qwen2} for identification, bge-reranker-v2-m3~\cite{chen2024bge} for re-ranking and LLama-3.1-8B-Instruct~\cite{grattafiori2024llama} for answer generation.
Hyper-parameters follow the default configurations provided by the authors and are summarized in Appendix~\ref{app:implement-details}.
For initial retrieval, we use a frozen EVA-CLIP-8B~\cite{sun2024eva} encoder to extract image features, following existing works~\cite{Echosight,ReflectiVA,Wiki-PRF}.
All image features are indexed and retrieved with cosine similarity using Faiss-GPU library~\cite{douze2025faiss}, consistent with prior practices~\cite{Echosight,ReflectiVA,Wiki-PRF}.

\subsection{Results on fixed datasets}
We first evaluate all methods on both the original and fixed versions of InfoSeek and E-VQA. 
Importantly, we ensure the evaluated samples from both original and fixed datasets preserve exactly the same sample IDs. We only correct flawed question formulations and annotated answers. Therefore, any performance difference reflects violations of \emph{Answer Derivability} and \emph{Question Well-Posedness} (Section~\ref{sec:issues}) rather than model changes.

\begin{table*}[!htbp]
\centering
\small
\caption{Performance comparison on InfoSeek and E-VQA before and after dataset fixing. For InfoSeek, most of methods performance improved, especially on Time and Numerical. \textbf{Bold} for best performance and {\ul underline} to the runner-up}
\label{tab:perf_comparison_unfixed_fixed}
\begin{tabular}{@{}llcccccccccc@{}}
\toprule
\multirow{3}{*}{Methods} & \multirow{3}{*}{Backbone} & \multicolumn{8}{c}{InfoSeek}                              & \multicolumn{2}{c}{E-VQA} \\ \cmidrule(l){3-12} 
                         &                           & \multicolumn{4}{c}{Unfixed} & \multicolumn{4}{c}{Fixed}   & Unfixed      & Fixed      \\ \cmidrule(lr){3-10}
                         &                           & All  & Time & Num  & String & All  & Time & Num  & String & All          & All        \\ \midrule

Wiki-PRF~\cite{Wiki-PRF}                 & Qwen                      & \textbf{44.9} & 35.1 & \textbf{45.7} & \textbf{45.3}   & \textbf{43.6} & {\ul 43.3} & 52.9 & \textbf{42.3}   & 31.9         & 33.1       \\
CoMEM~\cite{CoMeM}                    & Qwen                      & 24.3 & 16.5 & 19.3 & 25.5   & 23.5 & 17.5 & 21.1 & 24.3   & 14.1         & 13.4       \\
ReflectiVA~\cite{ReflectiVA}               & Llama                     & {\ul 37.3} & {\ul 37.1} & {\ul 43.9} & {\ul 36.4}   & 38.1 & 34.0 & {\ul 58.7} & 35.5   & 36.8         & 36.6       \\
IBA~\cite{IBA}                      & Llama                     & 34.5 & \textbf{38.1} & 43.0 & 33.0   & {\ul 42.4} & \textbf{47.4} & \textbf{61.9} & {\ul 39.4}   & \textbf{41.9}         & \textbf{42.7}       \\
EchoSight~\cite{Echosight}                & Llama                     & 30.9 & 29.9 & 38.1 & 30.0   & 37.2 & 32.0 & 51.1 & 35.6   & {\ul 40.4}         & {\ul 40.9}       \\
LLaVA-v1.5~\cite{liu2024improved}               & Llama                     & 5.8  & 0.0  & 7.2  & 6.0    & 6.6  & 1.0  & 9.0  & 6.6    & 12.9         & 12.7       \\
Qwen2.5-VL~\cite{bai2025qwen2}               & Qwen                      & 21.4 & 12.4 & 14.3 & 22.9   & 28.3 & 9.3  & 17.0 & 31.0   & 21.9         & 21.9       \\ \bottomrule
\end{tabular}
\end{table*}

By repairing QA annotations to enforce answer--evidence alignment and resolving underspecified questions, fixing yields substantial score changes across methods on InfoSeek.
Because the fixed and unfixed evaluations are aligned on the same sample IDs, these shifts reflect the impact of removing annotation- and template-induced confounders rather than changes in evaluation coverage.
Some methods improve markedly (\eg IBA $34.5 \rightarrow 42.4$, EchoSight $30.9 \rightarrow 37.2$), with larger gains on the stricter Time and Num subsets (\eg Wiki-PRF $35.1 \rightarrow 43.3$, $45.7 \rightarrow 52.9$).
Because images, retrieval settings, checkpoints, and metrics are held constant, these shifts isolate the impact of benchmark validity (Assumption~A/B) rather than changes in model design.
Crucially, we also observe ranking reversals after fixing, indicating that unfixed splits can distort comparative conclusions about which system components matter most.

\noindent \textbf{Key observations from fixing results.}
\textit{(i) Relative gaps shrink after fixing.}
On unfixed InfoSeek, aggregation/filtering pipelines appear much stronger than re-ranking based methods like IBA (\eg Wiki-PRF vs.\ IBA: $+10.4$; ReflectiVA vs.\ IBA: $+2.8$).
After fixing, these gaps shrink to $+1.2$ and $-4.3$, respectively.

\noindent \textit{(ii) Method rankings can reverse.}
On unfixed InfoSeek, ReflectiVA outperforms IBA, but the ordering reverses after fixing (IBA $42.4$ vs.\ ReflectiVA $38.1$).
On the Time subset, explicit evidence selection becomes more favorable after fixing (IBA $47.4$ and Wiki-PRF $43.3$ vs.\ ReflectiVA $34.0$).
These reversals matter in practice because they can change system-design conclusions.
For example, after the fixing, community may draw a conclusion that selecting the most relevant and informative evidence section is more effective than relying on model's inherent capability to assess the relevance of each evidence section to query.
Otherwise, if the community sticks to the conclusion draws from the unfixed data, a lot of efforts could be invested to expensive training but the gain may not aligned in a corresponding level. 
Similarly, the advantage of Wiki-PRF over IBA is large on unfixed InfoSeek ($44.9$ vs.\ $34.5$, gap $10.4$), but becomes much smaller after fixing ($43.6$ vs.\ $42.4$, gap $1.2$).
In this case, community may reconsider the trade-offs between developing new models and reconsidering the workflow orchestration.
Overall, violations of answer derivability and question well-posedness can distort comparative conclusions and misallocate optimization effort.

Due to space limits, we defer detailed fine-grained and qualitative analyses to Appendix~\ref{app:fixing-analysis} to show how our fixing improves the performance by correcting the desired annotation answer and improving the question to guide the desired answer.
On the shared grounded InfoSeek subset where EchoSight's top-1 reranked section belongs to the ground-truth entity (664 samples), question-only fixes improve QA accuracy from $60.1$ to $75.1$, joint question-answer fixes improve from $34.0$ to $45.2$, and answer-only fixes remain nearly unchanged. 
Detailed breakdowns are in Appendix~\ref{app:fixing-analysis}.
Post-retrieval results also provide supporting evidence for re-ranking methods that improving the question contributes to better re-ranking.
On InfoSeek, post-retrieval performance increases after fixing for EchoSight ($46.8 \rightarrow 48.9$) and IBA ($46.7 \rightarrow 47.8$).
This suggests that part of the QA improvement is associated with better evidence selection after fixing, while question and answer repairs likely contribute jointly.
Table~\ref{tab:fixing_stats} shows that InfoSeek has high modification rates for both questions and answers, whereas E-VQA mainly changes questions. This aligns with the observed pattern: E-VQA shifts are modest, while InfoSeek shows larger magnitude changes in Table~\ref{tab:perf_comparison_unfixed_fixed}.

\begin{table}[!htbp]
\small
\centering
\caption{Initial retrieval recall. Since initial retrieval are following the same image-to-image protocol of all baselines, the fixed and unfixed version share the same retrieved information. For Intra- and Inter-augmented version, augmented images are used.}
\label{tab:init_retrieval_recall}
\begin{tabular}{@{}lcccccccccc@{}}
\toprule
\multirow{2}{*}{Version} & \multicolumn{5}{c}{E-VQA}                               & \multicolumn{5}{c}{InfoSeek}                           \\ \cmidrule(l){2-6} \cmidrule(l){7-11} 
                         & R@1 & R@3 & R@5 & R@10 & R@20 & R@1 & R@3 & R@5 & R@10 & R@20 \\ \midrule
Full                     & 13.4     & 26.1     & 31.9     & 41.8      & 48.9      & 43.5     & 60.1     & 65.8     & 72.1      & 75.8      \\
Anchor                   & 12.8     & 26.3     & 32.6     & 43.3      & 50.7      & 43.1     & 59.9     & 66.1     & 72.7      & 76.4      \\
Intra-Aug              & 3.5      & 7.5      & 9.8      & 13.6      & 17.0        & 14.7     & 24.4     & 34.4     & 36.5      & 41.3      \\
Inter-Aug               & 2.9      & 6.7      & 8.6      & 12.3      & 15.0        & 20.4     & 29.5     & 31.9     & 34.2      & 34.4      \\ \bottomrule
\end{tabular}
\end{table}

\noindent \textbf{Retrieval performance affects cross-dataset rankings.}
We further report retrieval recall on the fixed full datasets and the anchor/augmented variants to analyze retrieval difficulty and to contextualize augmentation effects.
Table~\ref{tab:init_retrieval_recall} shows that InfoSeek has much higher initial R@1 than E-VQA (43.5\% vs 13.4\% on the full fixed set).
Correspondingly, Table~\ref{tab:post_retrieval_perf} shows that Wiki-PRF remains competitive on InfoSeek (48.6\%) but drops sharply on E-VQA (18.2\%), while EchoSight and IBA are more stable.
This pattern indicates that ranking differences across datasets are driven largely by retrieval difficulty rather than by downstream reasoning, which further strengthen the need to challenge the retrieval as we discussed in Section~\ref{sec:aug-pipeline}.

\begin{table}[!htbp]
\centering
\caption{Post-retrieval performance using retrieved \textbf{top 20} articles. on all dataset variants, including unfixed, fixed datasets, the subset of anchor samples and the augmented versions. Note for ReflectiVA\cite{ReflectiVA} and CoMEM\cite{CoMeM}, they aggregate the top-5 and top-10 articles \ie align with initial retrieval recall@5 and recall@10 in Table~\ref{tab:init_retrieval_recall}. }
\label{tab:post_retrieval_perf}
\begin{tabular}{@{}lcccccccccc@{}}
\toprule
\multirow{2}{*}{Method} & \multicolumn{5}{c}{E-VQA}                & \multicolumn{5}{c}{InfoSeek}             \\ \cmidrule(l){2-6} \cmidrule(l){7-11} 
                        & Unfixed & Fixed & Anchor & Intra & Inter & Unfixed & Fixed & Anchor & Intra & Inter \\ \midrule
EchoSight               & 36.4    & 36.3  & 46.6   & 11.8  & 11.7  & 46.8    & 48.9  & 48.4   & 16.8  & 20.4  \\
IBA                     & 34.7    & 35.0  & 44.8   & 8.9   & 9.1   & 46.7    & 47.8  & 45.1   & 21.7  & 21.6  \\
Wiki-PRF                & 18.0    & 18.2  & 17.9   & 4.9   & 4.7   & 48.2    & 48.6  & 48.3   & 18.1  & 17.6  \\ \bottomrule
\end{tabular}
\end{table}

\subsection{Results on augmented datasets}

To directly test the grounded-disambiguation requirement (Section~\ref{sec:aug-pipeline}), we evaluate on the augmented variants that introduce a single distractor while preserving the answer and KB.
We report three representative methods (EchoSight, IBA, Wiki-PRF) to cover re-ranking and aggregation paradigms.

\noindent \textbf{Assumption C (Grounded Disambiguation): augmentation removes the single-entity shortcut.}
Table~\ref{tab:init_retrieval_recall} shows that adding a single distractor sharply reduces initial retrieval recall.
On InfoSeek, R@1 drops from $43.5$ (Full) to $14.7/20.4$ (Intra/Inter); on E-VQA, it falls from $13.4$ to $3.5/2.9$.
Since KB and questions are unchanged, this collapse is attributable to the added entity ambiguity, confirming the augmentation effectively enforces grounded disambiguation.
To separate layout effects from semantic distractors, we additionally evaluate two layout-only controls: \textit{Blank} replaces the distractor with an empty panel and \textit{Double} duplicates the anchor image.
Initial retrieval is layout-sensitive, but Blank/Double remain easier than Intra/Inter (E-VQA R@1: $6.3/9.9$ vs.\ $3.5/2.9$ and InfoSeek R@1: $26.2/36.9$ vs.\ $14.7/20.4$).
Full initial- and post-retrieval results are reported in Appendix~\ref{app:augmentation-analysis}, showing that semantic distractors introduce difficulty beyond layout shift.

\noindent \textbf{Post-retrieval evidence selection does not recover the loss.}
Table~\ref{tab:post_retrieval_perf} shows that post-retrieval recall also degrades sharply on augmented images (\eg IBA InfoSeek $45.1 \rightarrow 21.7/21.6$, EchoSight E-VQA $46.6 \rightarrow 11.8/11.7$).
This suggests that once initial grounding fails, later evidence aggregation cannot compensate, aligning with our diagnosis that grounded disambiguation is the critical bottleneck.
{Surprisingly, the Wiki-PRF didn't try to actively invoke its grounding tool. We defer more detailed analysis in Appendix~\ref{app:augmentation-analysis}.}

\noindent \textbf{QA accuracy drops sharply, validating the lack of grounded disambiguation.}
Table~\ref{tab:QA-perf-augmented} reports QA accuracy on the anchor subset and its two augmented variants.
All methods suffer substantial performance degradation once ambiguity is introduced.
For example, on InfoSeek, IBA drops from $40.1$ (Anchor) to $21.4/21.6$ (Intra/Inter), and EchoSight drops from $38.7$ to $15.9/17.8$.
Even Wiki-PRF declines from $43.9$ to $23.6/25.9$.
Similar trends hold on E-VQA.

\begin{table*}[!htbp]
\caption{QA Performance on Intra-category(\textbf{Intra}) and Inter-Category(\textbf{Inter}) augmented datasets and the corresponding anchor samples. Following Table~\ref{tab:perf_comparison_unfixed_fixed}, we report the performance on E-VQA(\textbf{E-V})~\cite{EVQA} and InfoSeek(\textbf{IS})~\cite{InfoSeek} with breakdowns on Time(\textbf{T}), Numerical(\textbf{N}) and String(\textbf{S}) questions }
\label{tab:QA-perf-augmented}
\begin{tabular}{@{}lccccccccccccccc@{}}
\toprule
\multirow{3}{*}{Methods} & \multicolumn{5}{c}{Anchor}                             & \multicolumn{5}{c}{Intra}                        & \multicolumn{5}{c}{Inter}                        \\ \cmidrule(l){2-6} \cmidrule(l){7-11} \cmidrule(l){12-16} 
                         & \multicolumn{4}{c}{IS}   & \multirow{2}{*}{E-V} & \multicolumn{4}{c}{IS}   & \multirow{2}{*}{E-V} & \multicolumn{4}{c}{IS}   & \multirow{2}{*}{E-V} \\ \cmidrule(lr){2-5} \cmidrule(lr){7-10} \cmidrule(lr){12-15}
                         & All & T & N  & S &                       & All & T & N  & S &                       & All & T & N  & S &                       \\ \midrule
IBA                      & 40.1    & 45.8 & 25.9 & 41.7   & 38.4                  & 21.4    & 16.9 & 17.5 & 22.2   & 19.8                  & 21.6    & 28.9 & 14.8 & 22.1   & 16.6                  \\
Wiki-PRF                 & 43.9    & 33.7 & 1.6  & 50.5   & 32.8                  & 23.6    & 10.8 & 0    & 27.7   & 23.1                  & 25.9    & 22.9 & 0    & 29.9   & 22.5                  \\
EchoSight                & 38.7    & 30.1 & 22.8 & 41.4   & 42                    & 15.9    & 8.4  & 7.9  & 17.5   & 19.3                  & 17.8    & 14.5 & 9.5  & 19.1   & 18.5                  \\ \bottomrule
\end{tabular}
\end{table*}

These results indicate that, even after repairing QA validity (Section~\ref{sec:fixing-pipeline}), existing KB-VQA evaluations remain optimistic since their visual setups allow retrieval pipelines to exploit single-entity shortcuts.
Our controlled augmentation thus exposes a remaining bottleneck: \textbf{how to reliably ground the query to the correct entity and then select truly relevant evidence under clutter}, which suggests that future KB-VQA benchmarks should incorporate richer multi-entity interactions and stronger grounding requirements.

\section{Conclusion}\label{sec:Conclusions}

We show that current KB-VQA benchmarks can exhibit recurring dataset issues that complicate evaluation by answer accuracy alone.
Across InfoSeek and E-VQA, we identify three prominent confounders: (i) answer--evidence misalignment where annotated answers are not derivable from the benchmark-provided knowledge snapshot, (ii) underspecified questions that admit multiple plausible answers, and (iii) visually simplified settings that weaken the need for grounded disambiguation.
These issues can decouple scores from the knowledge-grounded retrieval and reasoning capability KB-VQA is intended to measure.

To address them, we propose two protocols.
First, a fixing protocol that enforces answer derivability and question well-posedness via evidence-aware auditing and targeted repair.
Second, a controlled augmentation strategy that introduces visual distractors while preserving the annotated answer, thereby increasing the need for grounded entity disambiguation under multi-entity conditions.
Together, these protocols transform existing benchmarks into a more diagnostic testbed for retrieval-and-reasoning.

Experiments on the repaired and augmented splits reveal trends that differ from standard evaluation.
After fixing, most methods improve substantially, and we observe that relative rankings can change, indicating that unfixed benchmarks may distort comparative conclusions and the perceived impact of system components.
Under augmentation, performance drops consistently for both retrieval recall and end-to-end accuracy, highlighting that robust grounding and intent-aligned retrieval remain challenging under visual ambiguity.

More broadly, our findings suggest that KB-VQA evaluation should explicitly account for benchmark validity.
We encourage the community to (i) prioritize evidence-derivable annotations and well-specified questions during dataset construction, (ii) report metrics that reflect evidence support and grounding robustness in addition to answer correctness, and (iii) develop retrieval and reasoning methods that remain reliable when multiple plausible entities and distractors are present.
We hope our protocols and revised benchmarks facilitate more faithful, diagnostic, and reproducible KB-VQA evaluation.

\section*{Acknowledgements}

This research was supported by the National Science Foundation (NSF) under grant numbers NSF2406647 and NSF2406648. It was also supported by the National Artificial Intelligence Research Resource (NAIRR) Pilot and the Delta advanced computing and data resource, which is supported by the National Science Foundation under award NSF-OAC-2005572.
S. M. R. and C. V. S. are supported by the Imageomics Institute, funded by the US National Science Foundation’s Harnessing the Data Revolution (HDR) program under Award No. 2118240 (Imageomics: A New Frontier of Biological Information Powered by Knowledge-Guided Machine Learning).

\bibliographystyle{splncs04}
\bibliography{main}

\newpage
\appendix
\pagenumbering{arabic}
\section{Qualitative Examples of Fixing Outcome}
\label{app:fixing-examples}
\begin{figure*}[!htbp]
    \centering
    \includegraphics[width=0.95\linewidth]{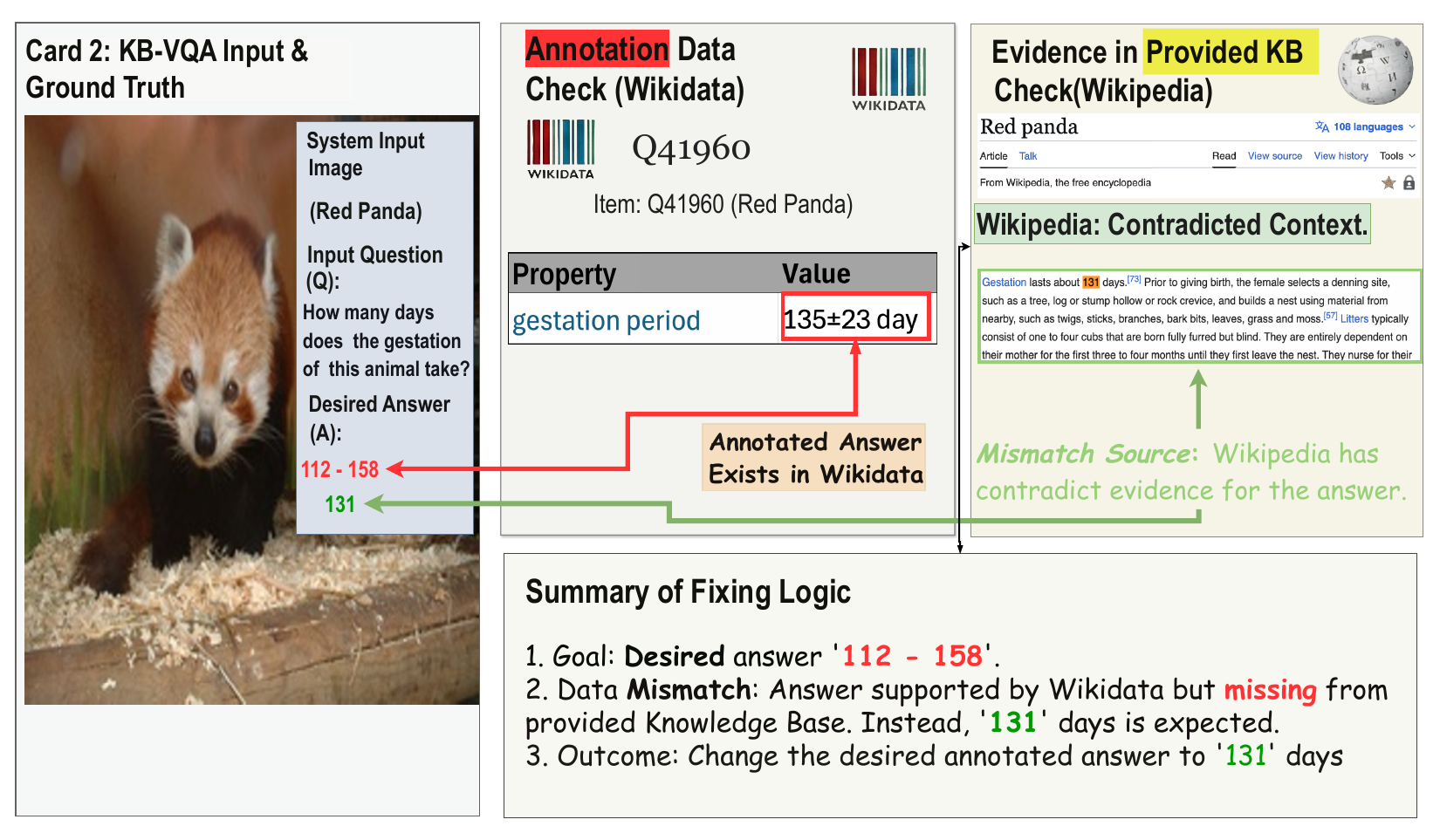}
    \vspace{-3mm}
    \caption{Fixing Qualitative example from InfoSeek (QID:9). InfoSeek~\cite{InfoSeek} selects answers from Wikidata~\cite{vrandevcic2014wikidata} triples and converts them into QA pairs, while the evaluation KB consists of Wikipedia articles. This cross-source construction can yield cases where the provided KB contains contradictory evidence to the annotated answer \eg, the desired answer is a range \textbf{`112 - 158'}, derived from $135 \pm 23$. Meanwhile the article in provided KB only contains \textbf{`131' days} . }
    \label{fig:fixing_outcome1}
\end{figure*}

\begin{figure*}[!htbp]
    \centering
    \includegraphics[width=0.95\linewidth]{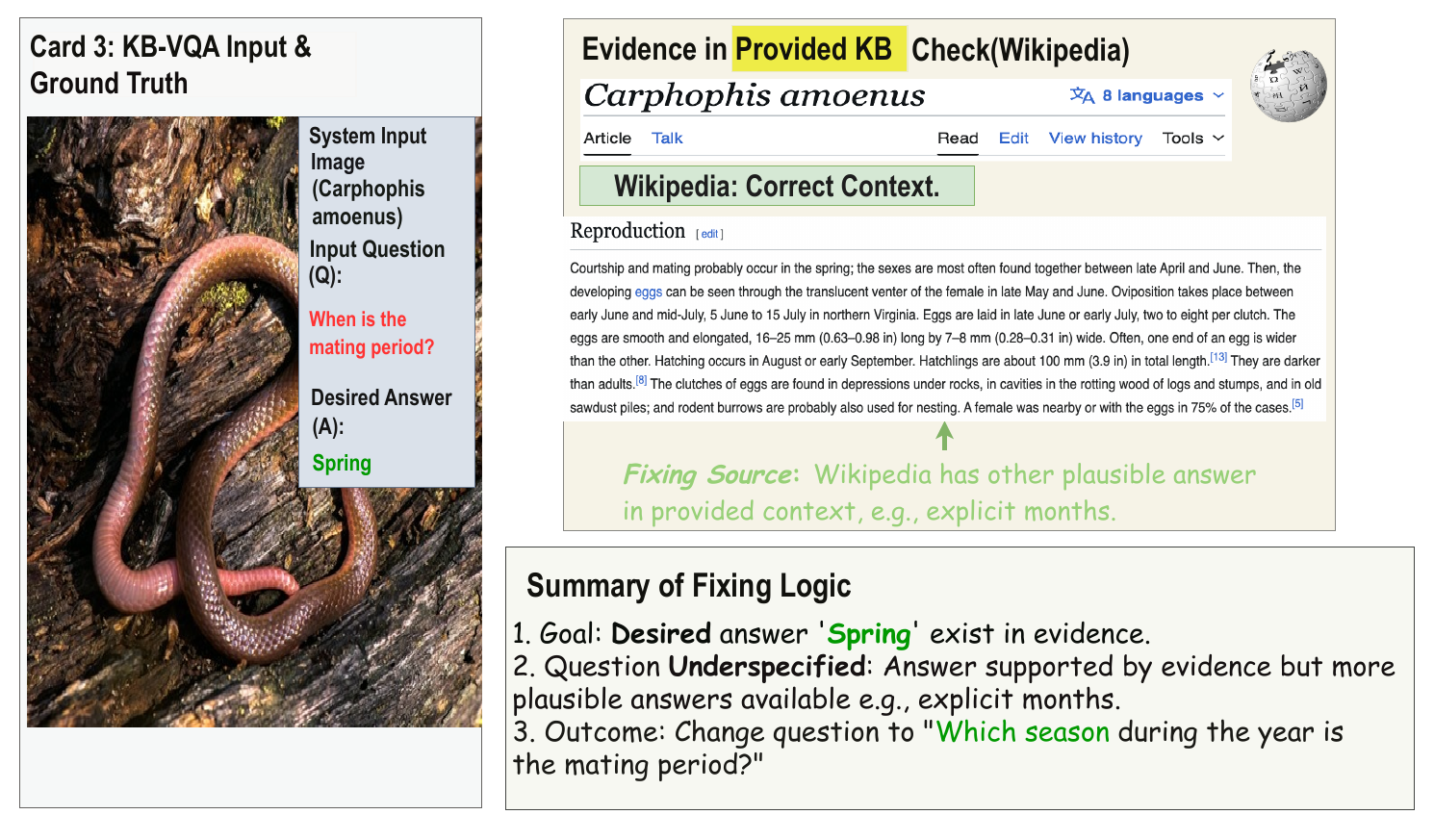}
    \vspace{-3mm}
    \caption{Fixing Qualitative example from E-VQA (QID:79) where desired answer `Spring' exists in evidence. Meanwhile there could be multiple plausible answers \eg, `April - June' since the following sentence indicates `are most often found between late April and June'.
    Therefore, the question is changed to \textbf{``Which season during the year is the mating period''}}
    \label{fig:fixing_outcome2}
\end{figure*}

In this section we present more qualitative samples of our fixing outcome.
Besides the qualitative example that is removed due to lack of supporting evidence as shown in Figure~\ref{fig:annotation_issues},
we also show how the instances are corrected to align with the supporting evidence in Figure~\ref{fig:fixing_outcome1} or improved with constraints in Figure~\ref{fig:fixing_outcome2} to ensure answer derivability and question well-posedness.

\section{Qualitative Examples of Augmentation Outcome}
\label{app:augmentation-examples}

\begin{figure}[!htbp]
    \centering
    \includegraphics[width=0.95\linewidth]{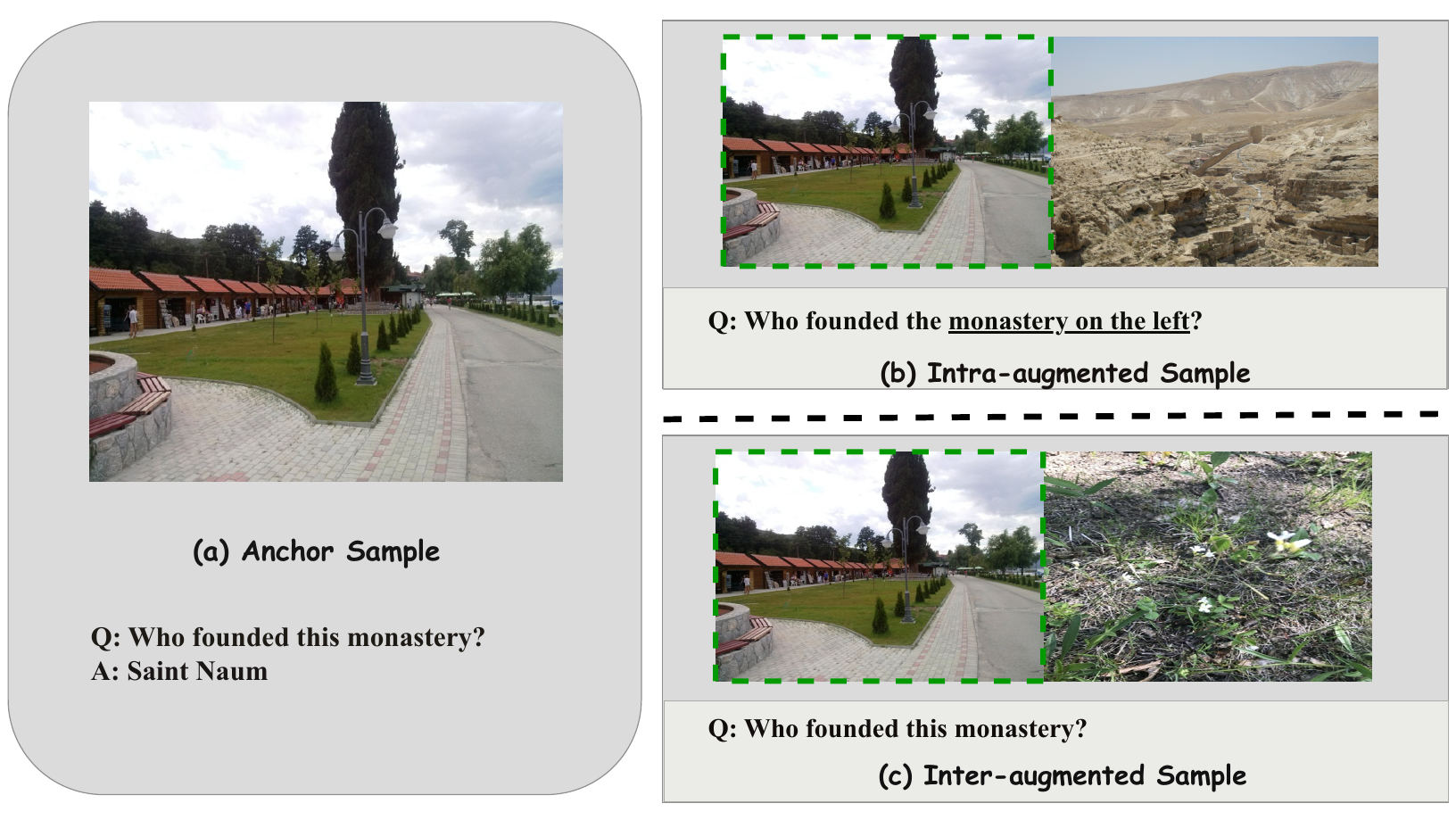}
    \caption{Qualitative example of our augmentation protocols (Section~\ref{sec:aug-pipeline}) on \emph{Monastery of Saint Naum}. All augmented variants keep the anchor answer in (a). (b) Intra-augmentation adds a minimal spatial cue (``on the left'') to preserve the original intent while inserting an additional monastery image (\emph{Mar Saba}) on the left. (c) Inter-augmentation inserts a visually distinct distractor (\emph{Arabidopsis lyrata}) while keeping the question unchanged.}
    \label{fig:augmentation-example2}
\end{figure}

\begin{figure}[!htbp]
    \centering
    \includegraphics[width=0.95\linewidth]{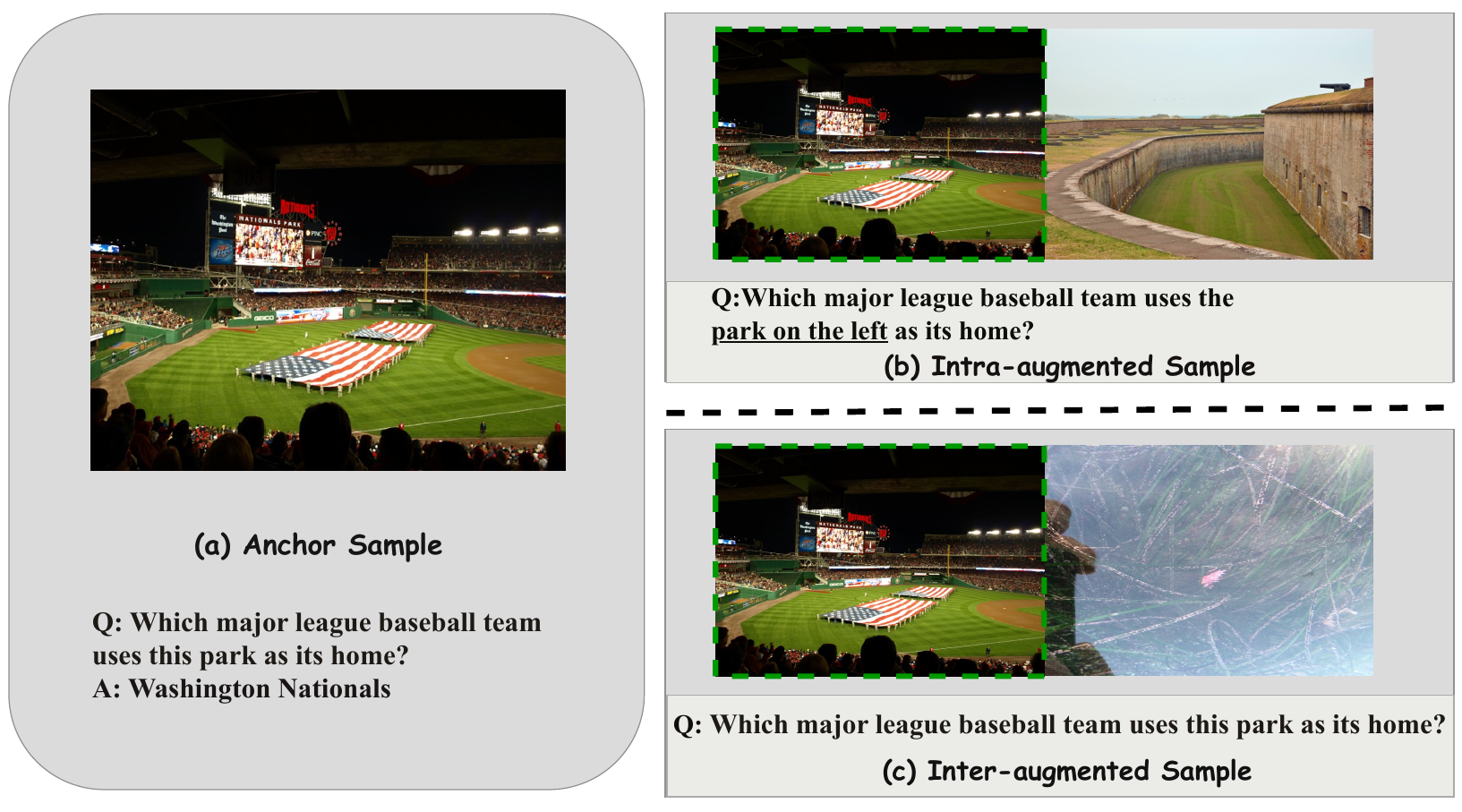}
    \caption{Qualitative example of our augmentation protocols (Section~\ref{sec:aug-pipeline}) on \emph{Nationals Park}. All augmented variants keep the anchor answer in (a). (b) Intra-augmentation adds a minimal spatial cue (``on the left'') to preserve the original intent while inserting an additional park image (\emph{Fort Macon State Park}) on the left. (c) Inter-augmentation inserts a visually distinct distractor (\emph{Okenia rosacea}) while keeping the question unchanged.}
    \label{fig:augmentation-example3}
\end{figure}

In this section, we provide more qualitative examples of our augmentation outcome, which is similar to Figure~\ref{fig:augmentation-example1}.
As shown in Figure~\ref{fig:augmentation-example2} and Figure~\ref{fig:augmentation-example3}, anchor images are augmented with either intra-category or inter-category entities with the augmentation protocol we propose in Section~\ref{sec:aug-pipeline}.

\section{Qualitative Examples of Error Cases}
\label{app:error-examples}
In this section, we provide some typical failure cases of benchmarked methods.
Basically, the failures result from errors in the initial retrieval, or from the re-ranking and filtering steps during the post-retrieval stage.

\begin{figure}[!htbp]
    \centering
    \includegraphics[width=0.95\linewidth]{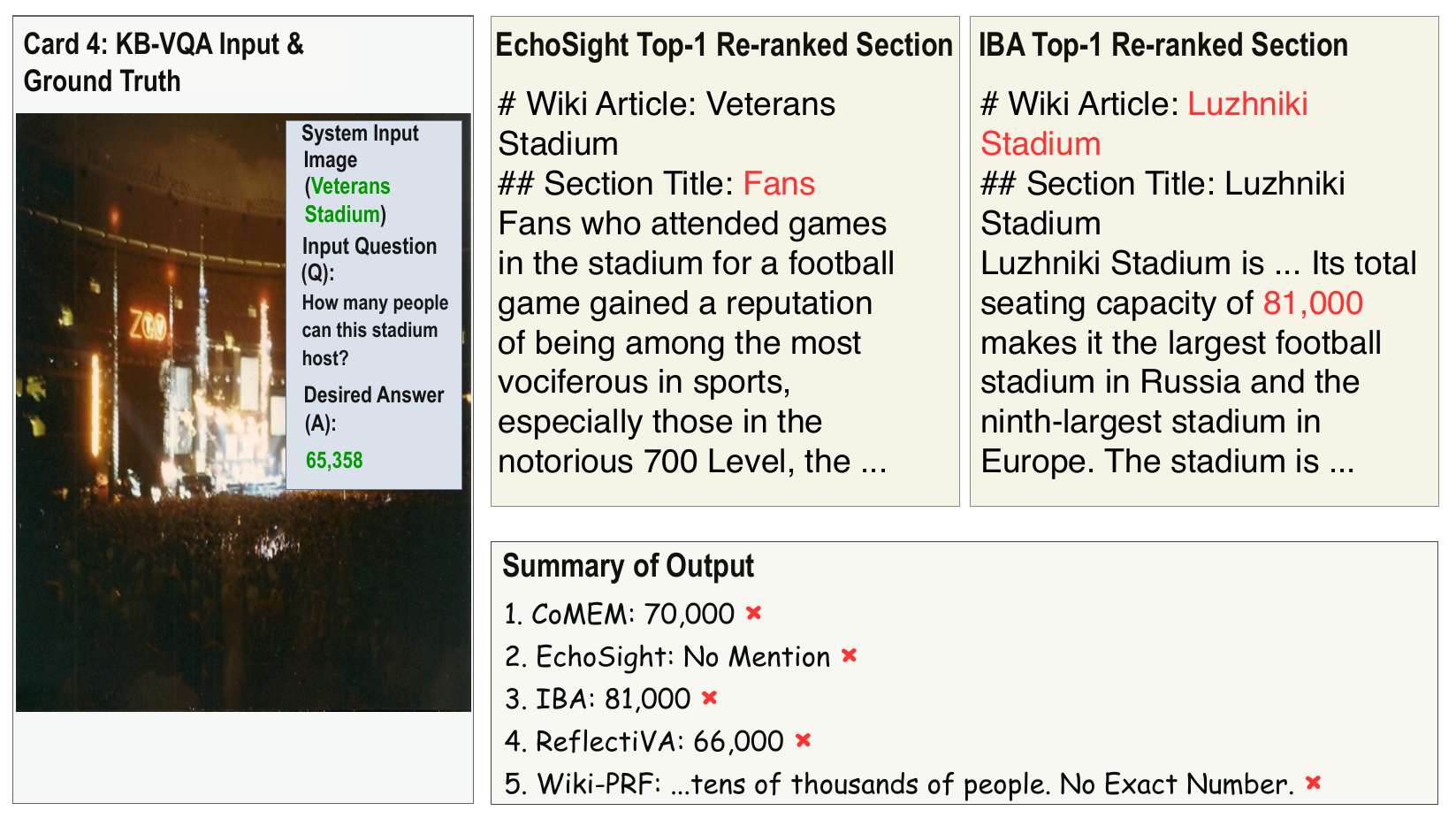}
    \caption{Qualitative example of failure cases on E-VQA~\cite{EVQA} \emph{Veterans Stadium}. The target entity is included in the initial retrieval results. However, all evaluated methods can't obtain the desired answer. EchoSight~\cite{Echosight} selects the section from the correct entity but it's irrelevant to the answer.
    IBA~\cite{IBA} selects a section from wrong entity. 
    The Aggregation/filtering methods, Wiki-PRF~\cite{Wiki-PRF}, CoMeM~\cite{CoMeM} and ReflectiVA~\cite{ReflectiVA} fail to derive the correct answer.}
    \label{fig:error-example1}
\end{figure}

When retrieval failed, no methods can ground their answer with correct evidence.
Hence, we provide the qualitative examples when initial retrieval succeeds, but the methods may also fail in re-ranking or filtering to correctly focus on the correct entity or sections on E-VQA. 
As shown in Figure~\ref{fig:error-example1},
IBA~\cite{IBA} picks a section from the wrong entity while EchoSight~\cite{Echosight} selects an irrelevant section from the target entity.

\section{Fine-grained analysis to fixed datasets}\label{app:fixing-analysis}

\begin{figure*}[!htbp]
    \centering
    \includegraphics[width=0.95\linewidth]{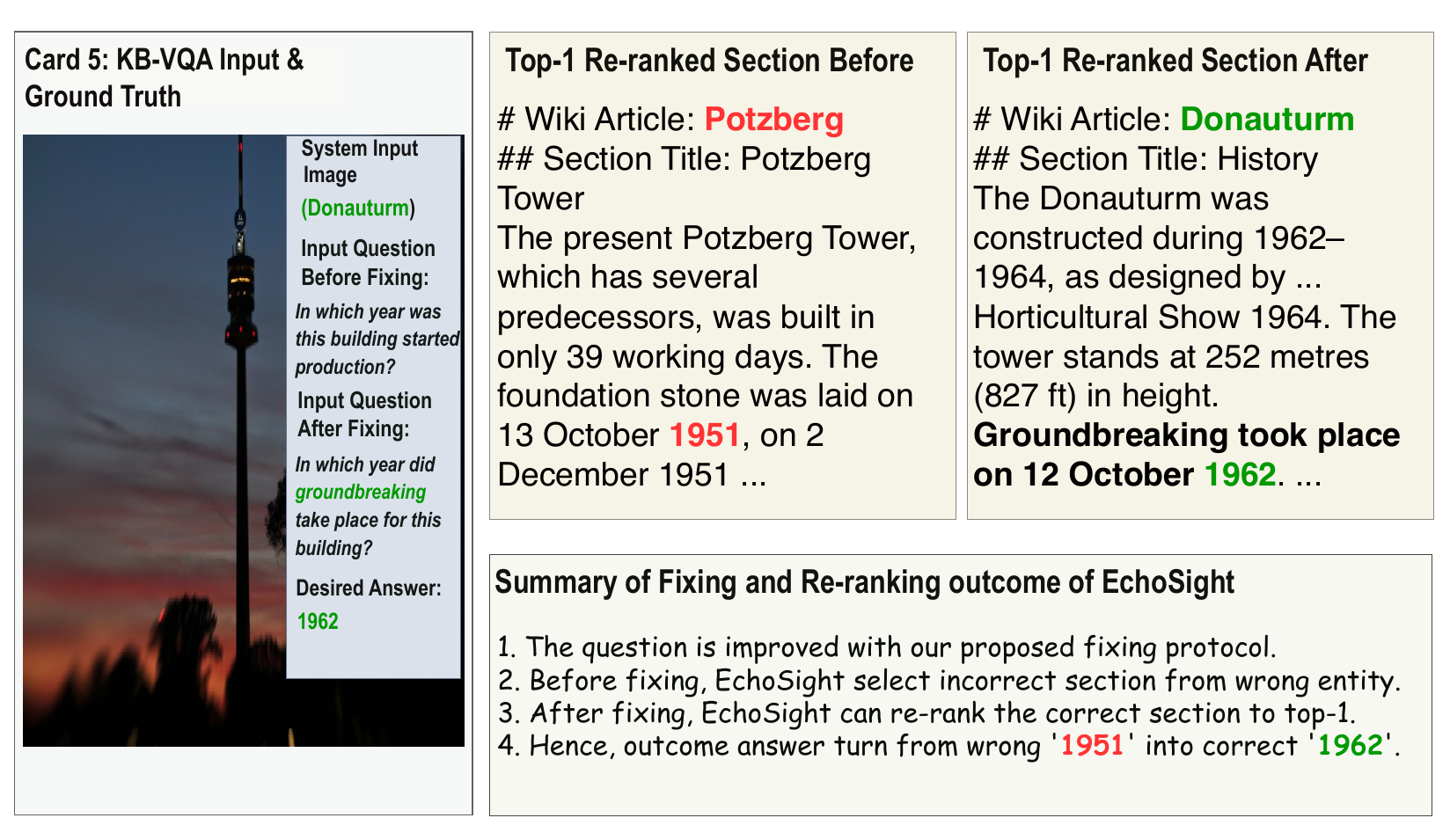}
    \vspace{-3mm}
    \caption{Fixing Qualitative example from InfoSeek (QID:40316) on \emph{Donauturm} . After fixing, the target section is correctly prioritized by EchoSight~\cite{Echosight} re-ranker. }
    \label{fig:fixing_improvement1}
\end{figure*}

In this section, we present some qualitative examples that fixing question and answer contributes to better re-ranking or answer generation.
As shown in Figure~\ref{fig:fixing_improvement1}, only revising the question under our proposed fixing protocols in Section~\ref{sec:fixing-pipeline} can better guide EchoSight~\cite{Echosight} to prioritize the correct section. 

\begin{figure*}[!htbp]
    \centering
    \includegraphics[width=0.95\linewidth]{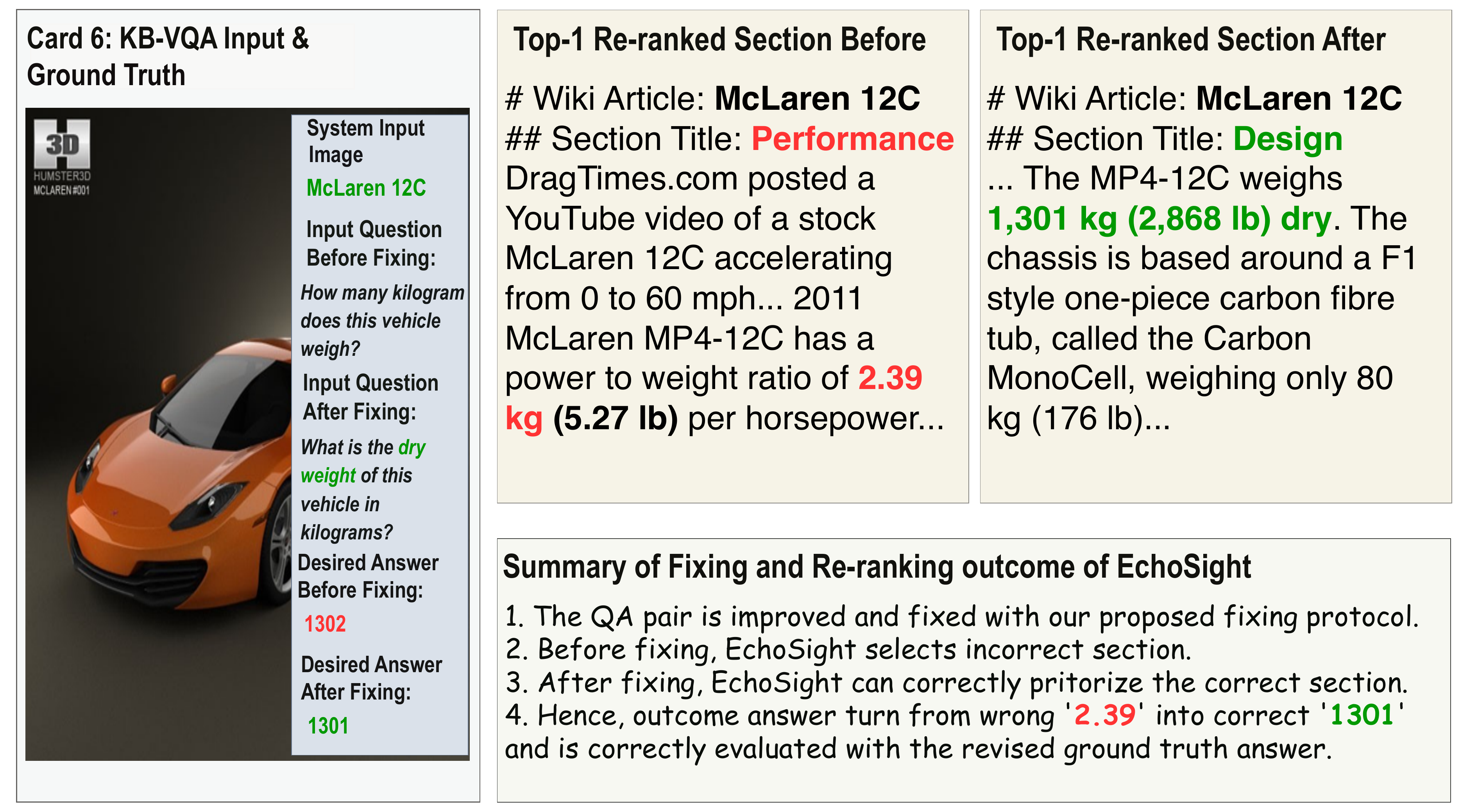}
    \vspace{-3mm}
    \caption{Fixing Qualitative example from InfoSeek (QID:5441) on \emph{McLaren 12C} . After fixing, the target section is correctly prioritized by EchoSight~\cite{Echosight} re-ranker. }
    \label{fig:fixing_improvement2}
\end{figure*}

In Figure~\ref{fig:fixing_improvement2}, with our proposed fixing protocol in Section~\ref{sec:fixing-pipeline}, we verify the absence of the desired answer \emph{`1302'} in provided knowledge base and revise the answer to \emph{`1301'}.
The question is also improved with proper constraint `dry weights'.
After the fixing, EchoSight~\cite{Echosight} can correctly prioritize the Design section and derive the desired answer \emph{`1301'} correctly.

We further analyze the shared grounded InfoSeek subset where EchoSight's top-1 reranked section belongs to the ground-truth entity.
As shown in Table~\ref{tab:fixing_grounded_subset}, question-only fixes are the main driver of improvement, while joint question-answer fixes also help and answer-only fixes are mostly unchanged.
For filtering, we manually check 100 removed InfoSeek samples and find that model accuracy remains 0 because the supporting evidence is absent from the benchmark-provided KB.

\begin{table}[!htbp]
\small
\centering
\caption{Fine-grained attribution on the shared grounded InfoSeek subset where EchoSight's top-1 reranked section belongs to the ground-truth entity.}
\label{tab:fixing_grounded_subset}
\begin{tabular}{lccc}
\toprule
Fix type & \# Samples & Before fixing & After fixing \\
\midrule
Question-only & 353 & 60.1 & 75.1 \\
Question-answer & 241 & 34.0 & 45.2 \\
Answer-only & 70 & 44.2 & 44.3 \\
\bottomrule
\end{tabular}
\end{table}

\begin{table}[!htbp]
\small
\centering
\caption{Fixing statistics on the entity-unique subset InfoSeek with 1924 samples and E-VQA with 4750 samples.}
\label{tab:fixing_stats}
\begin{tabular}{lcc}
\toprule
Dataset & Q changed & A changed \\
\midrule
E-VQA & 48.5\% & 2.9\% \\
InfoSeek & 76.1\% & 38.8\% \\
\bottomrule
\end{tabular}
\end{table}
\section{Fine-grained Analysis for augmentation}\label{app:augmentation-analysis}

In this section, we present more details on the performance of Wiki-PRF~\cite{Wiki-PRF} on the Intra-category and Inter-category augmented InfoSeek~\cite{InfoSeek} and E-VQA~\cite{EVQA} constructed with our augmentation protocols in Section~\ref{sec:aug-pipeline}.

As shown in Table~\ref{tab:InfoSeek-WikiPRF}, comparing with anchor samples, the numbers of tool calling and hit ratios of target entity of all tools drop significantly on both Intra-category and Inter-category augmented InfoSeek dataset.
Ideally, with more tool options, the retrieval performance of Wiki-PRF~\cite{Wiki-PRF} should be improved.
Specifically, with the grounding tool, it is expected that the distractors injected by our augmentation protocols in Section~\ref{sec:aug-pipeline} will be excluded.
However, it doesn't achieve better retrieval results.
We attribute this phenomenon to the nature of sparse rewards during its training process since it only focuses on the final answer accuracy without considering the potential reward in retrieval(\ie when retrieval hit the target entity.)
There are more calls on Intra-category augmented dataset than Inter-category but the ratio is lower.
We hypothesis phenomenon that VLM still struggles on comprehending spatial relations.

\begin{table}[!htbp]
\centering
\caption{Tool calling status of Wiki-PRF on the anchor, Intra-category and Inter-category augmented InfoSeek~\cite{InfoSeek} with 1,604 instances constructed with our augmentation protocols in Section~\ref{sec:aug-pipeline}.}
\label{tab:InfoSeek-WikiPRF}
\begin{tabular}{@{}lccccccccc@{}}
\toprule
\multicolumn{1}{c}{\multirow{2}{*}{Tool}} & \multicolumn{3}{c}{Number of Call} & \multicolumn{3}{c}{Hit@1} & \multicolumn{3}{c}{Hit@3} \\ \cmidrule(l){2-4} \cmidrule(l){5-7} \cmidrule(l){8-10} 
\multicolumn{1}{c}{}                      & Anchor     & Intra     & Inter     & Anchor  & Intra  & Inter  & Anchor  & Intra  & Inter  \\ \midrule
Caption                                   & 439        & 340       & 239       & 20.3    & 7.9    & 9.2    & 35.7    & 15.9   & 17.2   \\
Grounding                                 & 383        & 266       & 212       & 34.5    & 24.4   & 24.5   & 50.9    & 32.7   & 36.3   \\
Flipping                                 & 119        & 72        & 64        & 34.5    & 13.9   & 15.6   & 50.4    & 26.4   & 28.1   \\ \bottomrule
\end{tabular}
\end{table}

For Table~\ref{tab:EVQA-WikiPRF}, similar trend can be observed.
Both the numbers of tool calling and hit ratios of target entity of all tools drop significantly on both Intra-category and Inter-category augmented E-VQA dataset.
\begin{table}[!htbp]
\centering
\caption{Tool calling status of Wiki-PRF on the anchor, Intra-category and Inter-category augmented E-VQA~\cite{EVQA} with 3,871 instances constructed with our augmentation protocols in Section~\ref{sec:aug-pipeline}.}
\label{tab:EVQA-WikiPRF}
\begin{tabular}{@{}lccccccccc@{}}
\toprule
\multicolumn{1}{c}{\multirow{2}{*}{Tool}} & \multicolumn{3}{c}{Number of Call} & \multicolumn{3}{c}{Hit@1} & \multicolumn{3}{c}{Hit@3} \\ \cmidrule(l){2-4} \cmidrule(l){5-7} \cmidrule(l){8-10} 
\multicolumn{1}{c}{}                      & Anchor     & Intra     & Inter     & Anchor  & Intra  & Inter  & Anchor  & Intra  & Inter  \\ \midrule
Caption                                   & 794        & 717       & 580       & 7.2     & 1.4    & 1.7    & 10.3    & 2.4    & 3.6    \\
Grounding                                 & 867        & 805       & 530       & 11.2    & 4.9    & 6.4    & 16.2    & 7.7    & 9.4    \\
Flipping                                 & 207        & 311       & 98        & 17.9    & 2.6    & 3.1    & 27.1    & 3.5    & 5.1    \\ \bottomrule
\end{tabular}
\end{table}

We also include two layout-only controls to separate layout shift from semantic distractors: \textit{Blank} replaces the distractor with an empty panel, while \textit{Double} duplicates the anchor image.
Table~\ref{tab:blank_double_control} shows that initial retrieval is layout-sensitive, but Blank/Double remain easier than Intra/Inter and Wiki-PRF post-retrieval recovers much more under these layout-only controls.
This supports the conclusion that semantic distractors introduce additional grounded-disambiguation difficulty beyond layout shift.

\begin{table}[!htbp]
\small
\centering
\caption{Blank/Double layout-control results. Initial R@1 uses the same image-to-image retrieval protocol as Table~\ref{tab:init_retrieval_recall}; Wiki-PRF post-retrieval reports whether the final evidence pool contains the target entity.}
\label{tab:blank_double_control}
\begin{tabular}{lcccc}
\toprule
\multirow{2}{*}{Version} & \multicolumn{2}{c}{Initial R@1} & \multicolumn{2}{c}{Wiki-PRF post-retrieval} \\
\cmidrule(lr){2-3}\cmidrule(lr){4-5}
 & E-VQA & InfoSeek & E-VQA & InfoSeek \\
\midrule
Anchor & 12.8 & 43.1 & 17.9 & 48.3 \\
Blank & 6.3 & 26.2 & 13.0 & 37.0 \\
Double & 9.9 & 36.9 & 20.6 & 52.3 \\
Intra & 3.5 & 14.7 & 4.9 & 18.1 \\
Inter & 2.9 & 20.4 & 4.7 & 17.6 \\
\bottomrule
\end{tabular}
\end{table}
\section{Human Evaluation for Fixing Outcomes}
\label{app:fixing-human-eval}

Following prior dataset-quality evaluation practice~\cite{su2025skvqa}, we conduct human evaluation to further assess the quality of the fixed datasets. 
Specifically, we expand the review to 10\% of the fixed evaluation splits, including 475 E-VQA and 200 InfoSeek instances.
We hire two PhD student-level annotators to answer the VQA instances using the provided oracle evidence sections.
For each instance, annotators are instructed to answer the question solely based on the associated evidence section, and their responses are compared against the annotated ground-truth answers.

Overall, human evaluation shows strong agreement with the annotated answers, providing supporting evidence that the fixed datasets are of substantially improved quality under evaluation protocols adopted in prior work~\cite{su2025skvqa}. 
The mean accuracy $\pm$ standard deviation is $92.9 \pm 1.1$ for E-VQA and $91.5 \pm 2.1$ for InfoSeek.

\begin{figure}[!htbp]
    \centering
    \includegraphics[width=0.95\linewidth]{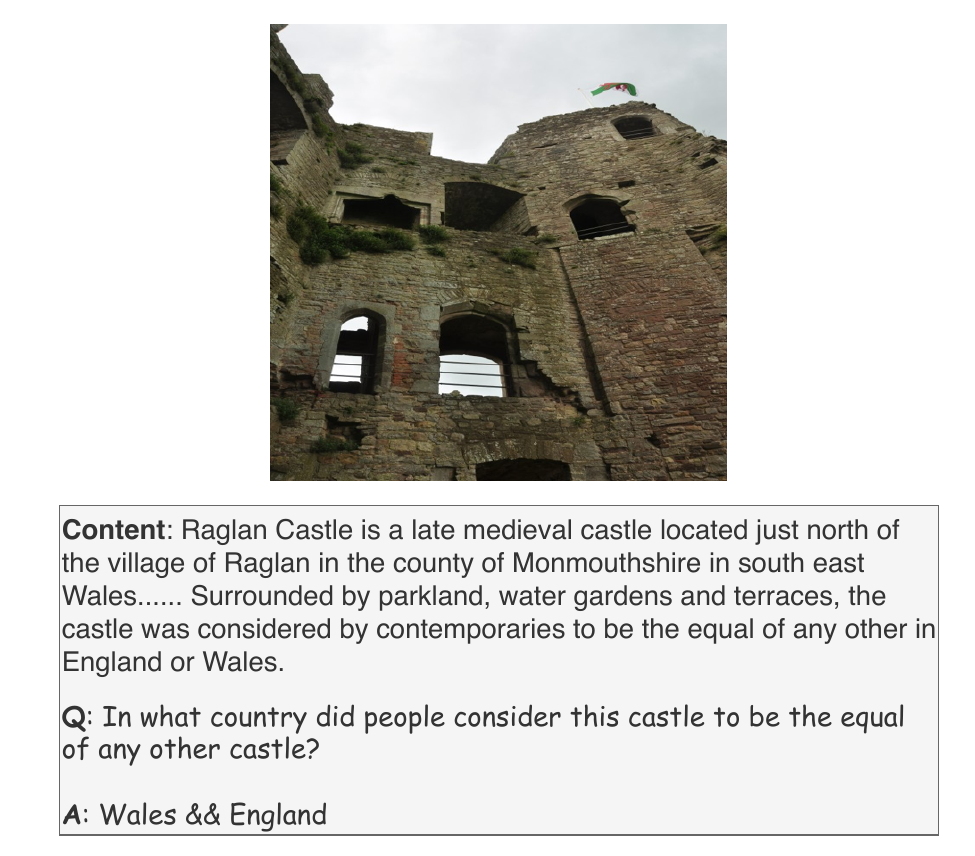}
    \caption{Qualitative example of human evaluation on \emph{Raglan Castle}. Human annotator has different understanding towards the subject of the question. According to evidence, the target entity is being compared with other castles located in England or Wales instead of people in Wales or England think this castle is comparable to other ones.}
    \label{fig:human-review-example2}
\end{figure}

\begin{figure}[!htbp]
    \centering
    \includegraphics[width=0.95\linewidth]{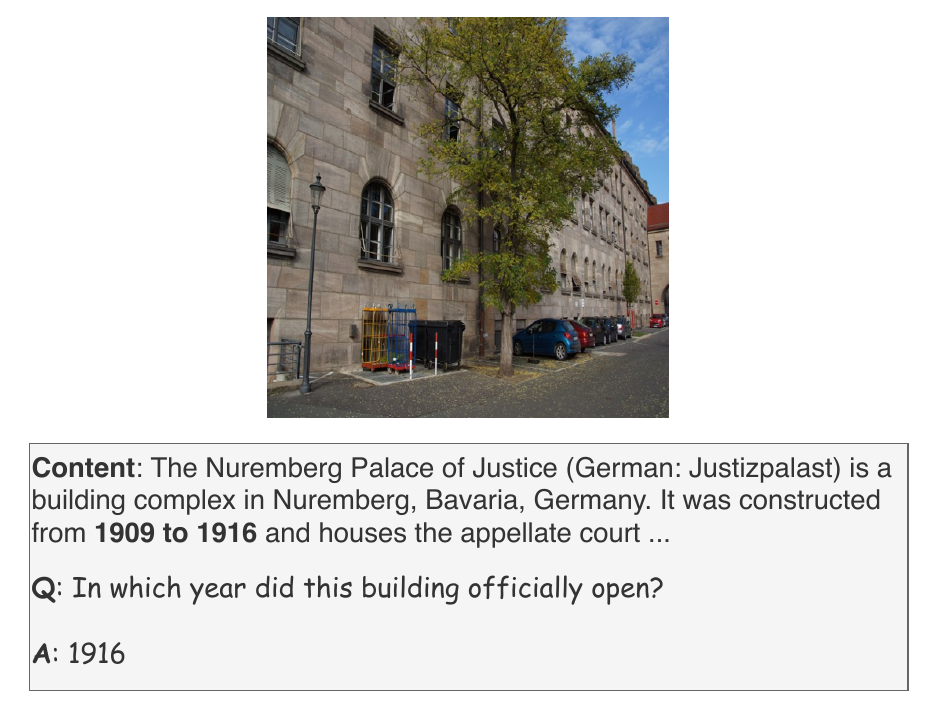}
    \caption{Qualitative example of human evaluation on \emph{Nuremberg palace of justice}. Construction accomplished in 1916 doesn't guarantee it will be open at the same time.}
    \label{fig:human-review-example3}
\end{figure}

However, despite this strong overall performance, a small number of cases still exhibit misalignment between human responses and the annotated answers. 
A closer examination shows that these failures are not merely random annotation noise. 
Instead, they reveal a harder category of residual issues that remain difficult to fix, particularly when the annotated answer is textually present in the evidence, yet the question itself remains ambiguous or semantically misaligned with the evidence as shown in Figure~\ref{fig:human-review-example2} and Figure~\ref{fig:human-review-example3}.

Figure~\ref{fig:human-review-example2} shows a case where the intended answer, `Wales and England', is supported by the evidence, yet the question is still ambiguous from a human perspective. 
The question, ``In what country did people consider this castle to be the equal of any other castle?'', can be interpreted in two ways. 
One interpretation asks in which country the relevant people were located, while another asks which countries' castles the target castle was said to equal. 
However, the supporting sentence only states that the castle ``was considered by contemporaries to be the equal of any other in England or Wales,'' which more naturally describes the comparison set of castles rather than the location of the people making the judgment. 
This example suggests that answer presence alone does not guarantee question well-posedness, and that some instances may still require additional rewriting to remove semantic ambiguity.

Figure~\ref{fig:human-review-example3} illustrates a different type of mismatch. 
The evidence states that the building was constructed from 1909 to 1916, while the question asks for the year in which it officially opened, with the annotated answer given as 1916. 
Human annotators noted that completion of construction does not necessarily imply the official opening date. 
Because the passage does not explicitly state the opening year, the annotated answer is not strictly derivable from the provided evidence. 
A semantically aligned revision would instead ask, for example, ``In which year was the construction of this building completed?''

These cases could be hard to define how a perfect fixing will be.
Although some of them may still be repairable, doing so would often require substantial rewriting of the question instead of minimal local edits.
This is not fully consistent with the minimal-change principle of our proposed fixing pipeline in Section~\ref{sec:fixing-pipeline}, since the meaning of the question may change significantly, for example, from asking when a building officially opened to when its construction was completed.
We leave such more aggressive remedies to future work. A possible direction is to use more and stronger proprietary models to determine when this kind of fundamental revision is necessary.

\section{Human Evaluation for Augmentation Outcomes}
\label{app:Augmentation-human-eval}

Following prior evaluation practice~\cite{su2025skvqa}, we further conduct human evaluation to assess the quality of the augmented datasets. 
Specifically, we hire two PhD-level annotators to answer the augmented VQA instances using the provided oracle evidence of the anchor entities. 
For each instance, annotators are instructed to answer the question based on the image and the associated evidence passage, and their responses are compared against the annotated ground-truth answers. 
When disagreement occurs, annotators are also asked to optionally report the main reason.

Overall, human evaluation suggests that the augmented datasets are of good quality. 
Human accuracy reaches $87.5\pm2.1\%$ and $96.0\pm1.4\%$ on E-VQA intra- and inter-augmentation, respectively, and $86.0\pm1.4\%$ and $89.5\pm3.5\%$ on InfoSeek intra- and inter-augmentation, respectively. 
These results provide supporting evidence that the augmented samples are generally answerable under evaluation protocols similar to those used in prior work~\cite{su2025skvqa}. 

At the same time, the remaining misaligned cases reveal several residual issues. 
Most of them are not caused by the answer being unsupported by the evidence, but rather by ambiguity in identifying the visual target after augmentation. 
In particular, the main challenge is that the original anchor image may already contain multiple plausible entities as in Figure~\ref{fig:augmentation-example3}, or the added distractor image may introduce visually confusing foreground or background content as in Figure~\ref{fig:augmentation-example3}. 
These issues may still be further improved, but doing so would likely require much stronger image-specific models or operations, such as more reliable object detection, grounding, or image cropping to accurately locate the target entity.

\begin{figure}[!htbp]
    \centering
    \includegraphics[width=0.95\linewidth]{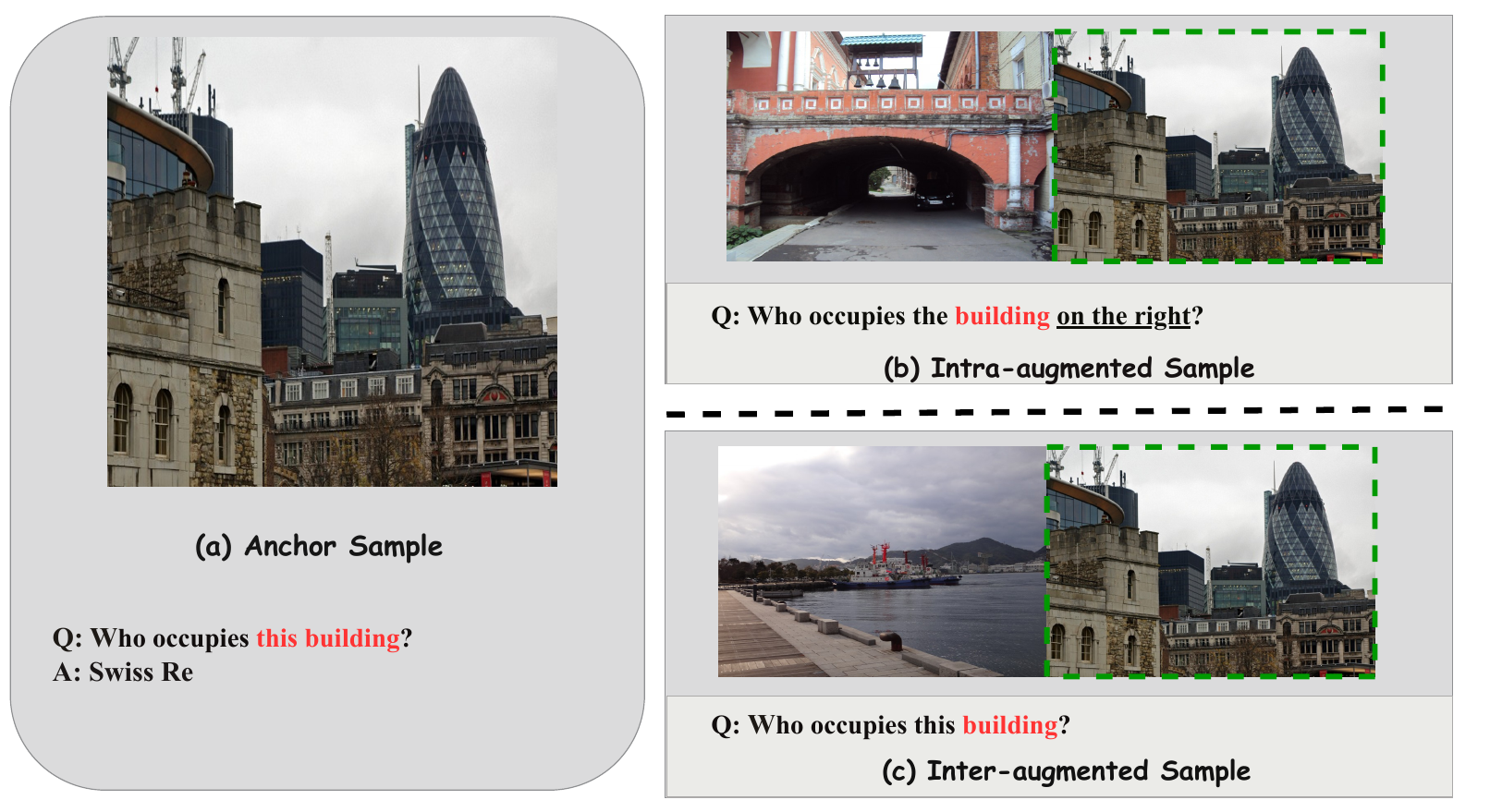}
    \caption{Qualitative example of human evaluation on augmented examples on \emph{30 St Mary Axe}. However, multiple buildings are shown in the query image of original anchor question. Hence in the augmented images, the target entity is still unclear. The annotators suggest further improve the question \eg, `Who occupies the skyscraper?'}
    \label{fig:human-review-augmentation-example3}
\end{figure}

Figure~\ref{fig:human-review-augmentation-example3} shows a representative case where the ambiguity already exists in the original anchor image. 
In the anchor example on \emph{30 St Mary Axe}, multiple buildings are visible, and the question asks, ``Who occupies this building?'' 
After augmentation, the question remains difficult because the target building is still not uniquely specified. 
Even in the intra-augmented case, where the phrase ``on the right'' is added, annotators noted that the right side still contains multiple plausible buildings, including both a lower building and the more salient skyscraper. 
This suggests that spatial cues alone may not always be sufficient when the anchor image itself contains multiple visually plausible targets. 
A possible improvement would be to use stronger grounding or cropping methods to isolate the intended entity more clearly, or to further refine the question wording (\eg , ``Who occupies the skyscraper?'').

\begin{figure}[!htbp]
    \centering
    \includegraphics[width=0.95\linewidth]{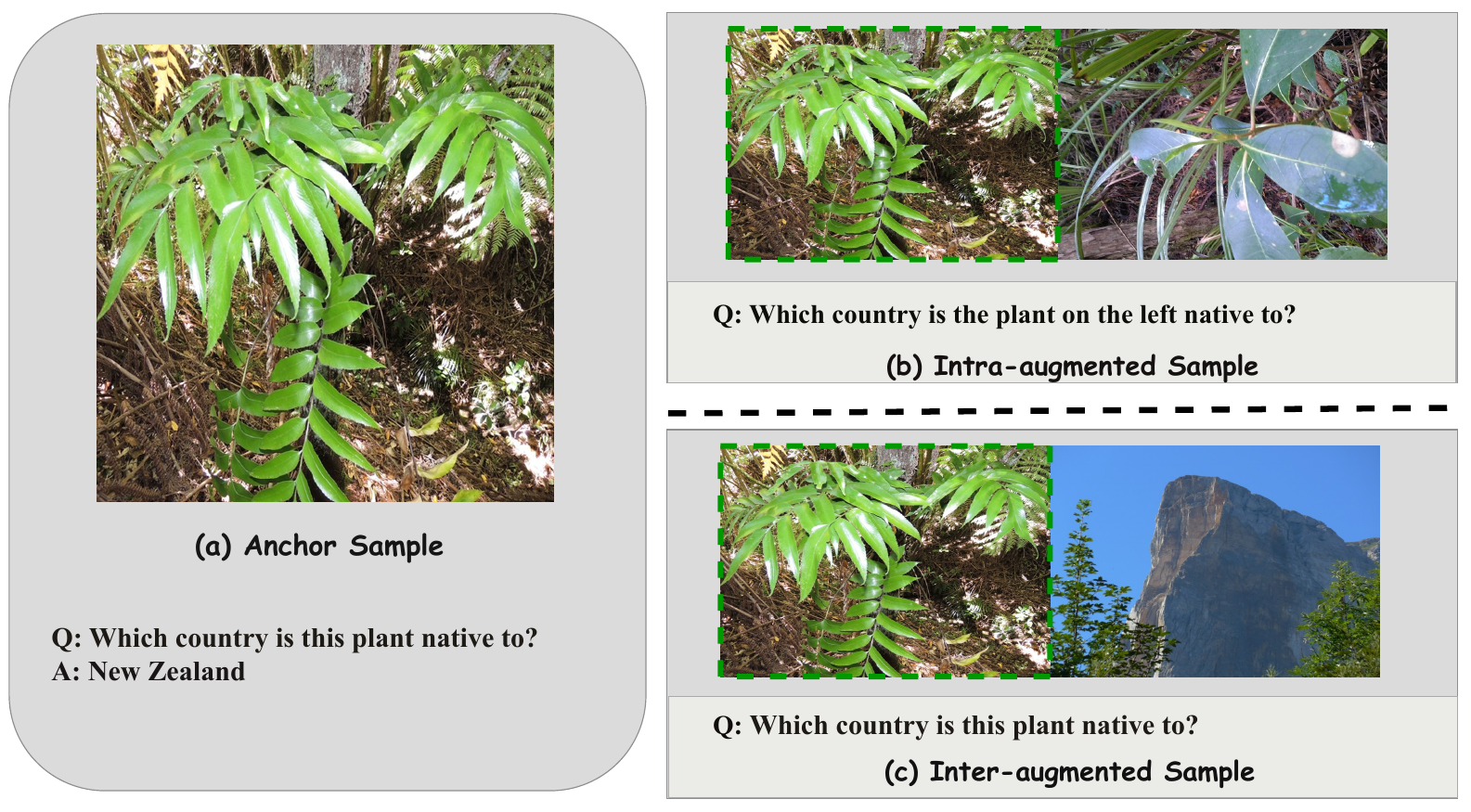}
    \caption{Qualitative example of human evaluation on augmented examples on \emph{Asplenium oblongifolium}. 
    For intra-augmented, the distractor is an image of \emph{Coprosma lucida}.
    Since the location `On the left' is provided, the question is still clear.
    However,  for inter-augmented, there are trees for the distractor image of \emph{Mont Aiguille}. Therefore, without further instructions, it will be hard to decide the target entity.}
    \label{fig:human-review-augmentation-example2}
\end{figure}

We also observe cases where the distractor image introduces additional foreground or background entities that interfere with target identification. 
Figure~\ref{fig:human-review-augmentation-example2} gives one such example on \emph{Asplenium oblongifolium}. 
In the intra-augmented case, the distractor image is \emph{Coprosma lucida}, and the spatial cue ``on the left'' is sufficient to keep the question clear. 
However, in the inter-augmented case, the distractor image of \emph{Mont Aiguille} also contains visible trees and vegetation. 
As a result, annotators reported that it becomes less clear which plant the question refers to. 
This example shows that inter-category distractors can still accidentally introduce same-type visual content, making the target harder to identify without additional grounding signals. 
A possible remedy is to use more careful distractor filtering or stronger image manipulation tools to suppress irrelevant visual entities.

\begin{figure}[!htbp]
    \centering
    \includegraphics[width=0.95\linewidth]{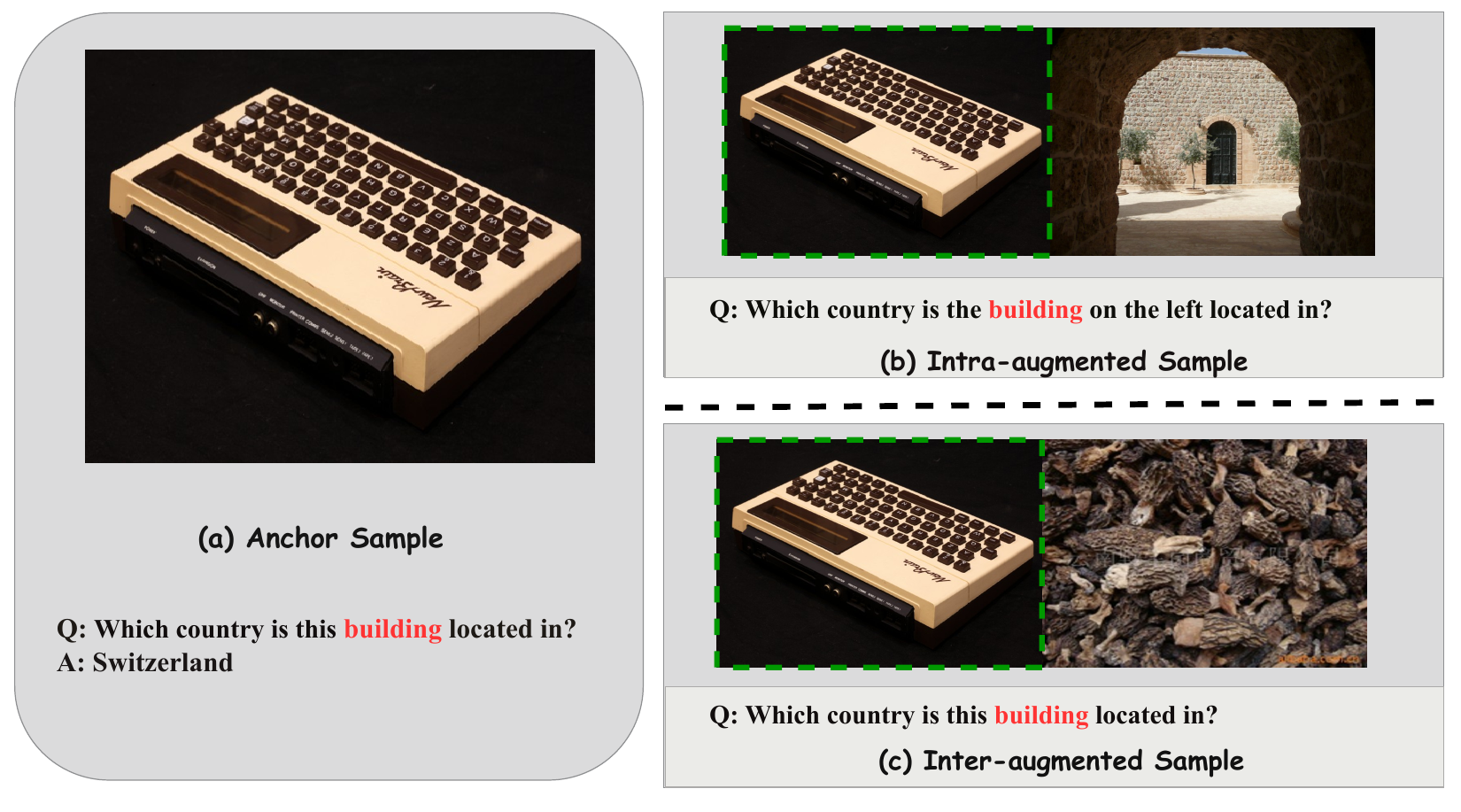}
    \caption{ In a very rare case, we observe that there is a misalignment between query image and text on subject. Qualitative example of human evaluation on augmented examples on \emph{Musée Bolo}. The query image in original anchor question is a computer collected in the museum instead of the building itself. Hence in the augmented images, the target entity could be ambiguous. For example in the intra-augmented image, even on the left is indicated, the visually available building is only the `Mor Gabriel Monastery', which is introduced as a distractor.}
    \label{fig:human-review-augmentation-example1}
\end{figure}

Meanwhile, in very rare cases(only 2 for the 100 sampled InfoSeek augmented instances ), we observe a more fundamental issue that the original anchor image itself does not clearly match the textual subject. 
As shown in Figure~\ref{fig:human-review-augmentation-example1}, the example on \emph{Musée Bolo} uses an image of a computer exhibited in the museum rather than the museum building itself, while the question asks, ``Which country is this building located in?'' 
This creates a mismatch between the query image and the target subject, which then carries over into the augmented versions \eg, even asking with spatial cues, the visually available building is only the distractor entity, which may disobey the original intention. 
Addressing this type of problem may require an additional verification stage with powerful closed proprietary VLMs to ensure that the anchor image is visually aligned with the textual subject.

Overall, these cases could be hard to define how a flawless fixing will be.
They may still be fixable by removing the irrelevant entities in foreground or background or locating the intended object, but doing so would often require stronger image-specific processing with specific models and supervision.
These remedies may go beyond the scope of our current augmentation pipeline in Section~\ref{sec:aug-pipeline} to enforce grounded disambiguation and therefore could be left to future work. 
A possible direction is to use more powerful multi-modal models or specialized vision modules to decide when such stronger intervention is necessary and how the visual input should be revised.
\section{Implementation Details}\label{app:implement-details}

For EchoSight~\cite{Echosight}, we use the provided code and re-ranker checkpoint to conduct the experiments, the answer generator we use is LLaMA-3.1-8B-Instruct~\cite{grattafiori2024llama}.
For IBA~\cite{IBA}, we use Qwen2.5-VL-7B-Instruct to narrow down the entity scope from the 20 initially retrieved entities to 3 and use bge-v2-m3 textual re-ranker.
For ReflectiVA and CoMEM, we use top-5 and top-10 retrieved articles to conduct experiments.
For Wiki-PRF~\cite{Wiki-PRF}, we use the released checkpoint and code to conduct experiments, the post-retrieval content consists of top-1 initial visual retrieval result and top-3 re-ranked sections via tool calling retrieval(\eg, grounding or captioning). 
Following paper's design, the post-retrieved content will be further filtered to support final answer generation.
\end{document}